%% file: main.tex
\documentclass[10pt]{article}
\usepackage{graphicx}
\usepackage{bm}
\usepackage{psfrag}
\usepackage{subfigure}
\usepackage{cite}
\usepackage{hyperref}
\usepackage[normalem]{ulem}
\usepackage{pgf}
\usepackage{pgfplots}
\usepackage{tikz}
\usetikzlibrary{decorations.pathreplacing}
\usepackage{import}
\usepackage{wrapfig}
\usepackage{pdfpages}
\usepackage{nicefrac}
\usepackage{amsmath}
\usepackage{amssymb}
\usepackage{mathtools}
\usepackage{microtype}
\usepackage{lipsum}
\usepackage{authblk}

\newcommand{\be}{\begin{equation}}
\newcommand{\ee}{\end{equation}}
\newcommand{\ba}{\begin{eqnarray}}
\newcommand{\ea}{\end{eqnarray}}

\title{Perspective: A Phase Diagram for Deep Learning unifying Jamming, Feature Learning and Lazy Training}
\author[a]{Mario Geiger}
\author[a]{Leonardo Petrini}
\author[a]{Matthieu Wyart}
\affil[a]{Institute of Physics, \'Ecole Polytechnique F\'ed\'erale de Lausanne, 1015 Lausanne, Switzerland}

\begin{document} 
\date{}

\maketitle

\begin{abstract}
Deep learning algorithms are responsible for a technological revolution in a variety of tasks including image recognition or Go playing. Yet, why they work is not understood. Ultimately, they manage to classify data lying in high dimension -- a feat generically impossible due to the geometry of high dimensional space and the associated {\it curse of dimensionality}. Understanding what kind of structure, symmetry or invariance makes data such as images learnable is a fundamental challenge. Other puzzles include that (i) learning corresponds to minimizing a loss in high dimension, which is in general not convex and could well get stuck bad minima. (ii) Deep learning predicting power increases with the number of fitting parameters, even in a regime where data are perfectly fitted. In this manuscript, we review recent results elucidating (i,ii) and the perspective they offer on the (still unexplained) curse of dimensionality paradox. We base our theoretical discussion on the $(h,\alpha)$ plane where $h$ is the network width and $\alpha$ the scale of the output of the network at initialization, and provide new systematic measures of performance in that plane for MNIST and CIFAR 10. We argue that different learning regimes can be organized into a phase diagram. A line of critical points sharply delimits an under-parametrised phase from an over-parametrized one. In over-parametrized nets, learning can operate in two regimes separated by a smooth cross-over. At large initialization, it corresponds to a kernel method, whereas for small initializations features can be learnt, together with invariants in the data. We review the properties of these different phases, of the transition separating them and some open questions. Our treatment emphasizes analogies with physical systems, scaling arguments and the development of numerical observables to quantitatively test these results empirically. Practical implications  are also discussed, including the benefit of averaging nets with distinct initial weights, or the choice of parameters $(h,\alpha)$ optimizing performance.
\end{abstract}

\section{Introduction}

One of the prerequisites of human or artificial intelligence is to make sense of data  that often lie in large dimension.
A classical case is computer vision, where one seeks to classify the content of a picture \cite{marr1979computational} whose  dimension  is  
the number of pixels. In the supervised setting considered in this review,   algorithms are trained  from known, labeled data. 
For example, one is given a training set of one million pictures of cats and  dogs, and knows which is which. The goal is to  build an algorithm that can learn a rule from these examples, and predict
if a  new picture presents a cat or a dog. After sixty years of rather moderate progress, machine learning is undergoing a revolution.
Deep learning algorithms \cite{Lecun15}, which are inspired from the organisation of our visual cortex, are now remarkably successful at a wide range of tasks including
speech \cite{amodei2016deep} and text recognition \cite{shi2016end}, self-driving cars \cite{huval2015empirical} and  beating the best humans at Go  \cite{silver2017mastering} or video games \cite{mnih2013playing}. Yet, there is very limited understanding as to why these algorithms work. 
As explained below, there are several reasons why deep learning in particular and supervised learning in general should not work. Most prominently, the {\it curse of dimensionality}
associated with the geometry of space in large dimension  prohibits learning  in a generic setting. If high dimensional data can be learned, then there must have a lot of structure, invariances 
and symmetries. Understanding what is the nature of this structure and how it can be harvested by neural nets with suitable architectures is a  challenge of our times.

In the part \ref{intro1} of this introduction,  we  review the basic procedure of supervised learning with deep nets, together with the quantification of its success via a learning curve exponent. In part  \ref{intro2}, we discuss several reasons making this success surprising.
It includes  the curse of dimensionality mentioned above,  the putative presence of the bad minima in the loss landscape and  the fact that deep learning typically works in an over-parametrized regime where the number of fitting parameters is much larger than the number of data.
In part  \ref{intro3}, we review two recent ideas seeking to address these paradoxes. First, in the limit where the network width (and thus the number of parameters) diverges, deep learning  converges to two distinct, well-defined algorithms ({\it lazy training} and {\it feature learning}) depending on the scale of initialization. In each regime, a global minimum of the loss is found. Second, in the context of image classification it was hypothesized that deep learning  efficiently deals with the curse of dimensionality  because neural nets learn to become insensitive to smooth deformations of the image. 

These two body of work raise various questions, detailed below. In a nutshell: (i)  Is there a critical number of parameters where one enters in the over-parametrized regime and bad minima in the loss disappear? What is the nature of this transition, and the geometry of the loss-landscape in its vicinity? (ii) Why does performance tends to improve asymptotically with the number of parameters even past this transition? (iii) Once in the over-parametrized regime, where does the cross-over between lazy training and feature learning takes place? Which regime  performs better, and how does it depend on the data structure and network architecture? (iv) How are simple invariants  learnt in the feature learning regime, and how does it affect the learning curve exponent? The goal of this manuscript, laid out in the part \ref{intro4} of this introduction, is both to review recent results addressing these questions in the context of classification in deep nets, as well as to provide new systematic empirical data unifying these results into a phase diagram.

\subsection{Presentation of deep learning and quantification of its successes}
\label{intro1}
\paragraph{Signal propagation in some architectures:}
Deep learning is a fitting procedure, in which the functional form used to interpolate the data depends on many parameters, and can be represented iteratively. 
State-of-the-art (SOTA) architectures characterising this functional form can be extremely complex, here we restrict ourselves to  essential properties.   We denote by $\tilde f_{\bf \theta}(\mathbf{x})$ the output of a network corresponding to an input $\mathbf{x}$, parametrized by $\mathbf{\theta}$. We reserve the notation $f_{\bf \theta}(\mathbf{x})$ for the prediction model, the relation between the two will be defined in Eq.\ref{eq:bachmodel}.
\begin{figure}[t]
    \centering
    \setlength{\unitlength}{0.1\textwidth}
    \includegraphics[width=\textwidth]{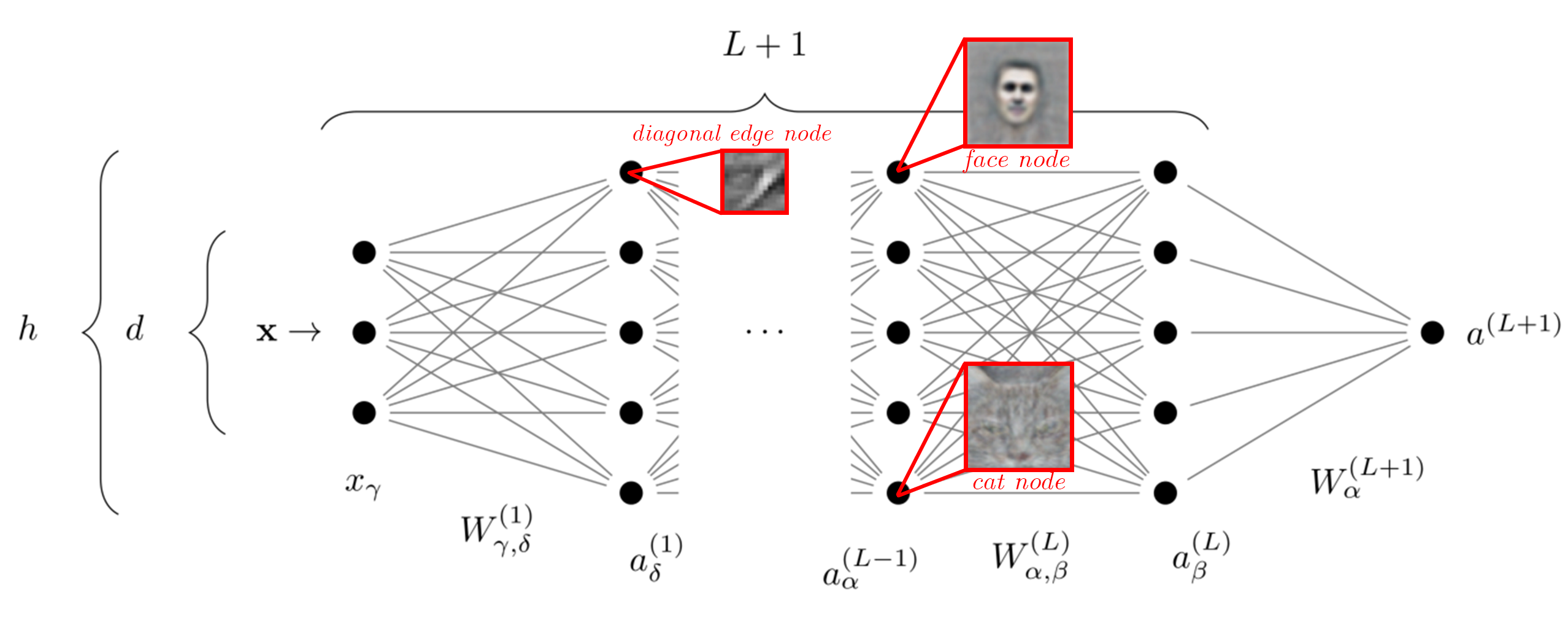}
    \caption{Architecture of a fully-connected network with $L$ hidden layers of neurons of constant size $h$.
    Points indicate neurons, connections between them are characterized by a weight. Biases are not represented here. The $W$'s are the weights, $\mathbf{x}$ is the input, and the $a$'s are the preactivations as defined in Equations (1-3). Taken from \cite{Geiger18}.
    \underline{In red}: Neural representation of the data at different depths (illustrative). Neurons in the first layers respond to local, simple features, such as edges. 
     Deeper in the network, neurons respond to more and more high-level features such as cats or human faces. Adapted from \cite{le2013building}.
     \label{fig:architecture}}
\end{figure}
For fully connected nets (FC) -- which, from a theoretical point of view, are the most studied architectures -- 
the output function can be written recursively as:
\begin{eqnarray}
    \tilde f_{\bf \theta}(\mathbf{x}) \equiv a^{(L+1)},\vspace{0.5em}\\
    a^{(i)}_\beta = \sum_\alpha \frac{1}{\sqrt{h}} W^{(i)}_{\alpha,\beta}\,\sigma\left(a^{(i-1)}_\alpha\right) + B^{(i)}_\beta,\\
    a^{(1)}_\beta = \sum_\alpha \frac{1}{\sqrt{d}} W^{(1)}_{\alpha,\beta}\,x_\alpha + B^{(1)}_\beta,
    \label{eq:recnnet}
\end{eqnarray}
where $a^{(i)}_\alpha$ is the preactivation of  neuron $\alpha$, located at depth $i$ and $L$ the total number of layers. We consider networks with a fixed number of hidden layers $L$, each of size $h$. In the following, given that the total number of parameters follows $N \sim Lh^2$, we'll use both $h$ and $N$ to refer to the network size.
In our notation the set of parameters $\mathbf{\theta}$ includes both the weights $W^{(i)}_{\alpha,\beta}$ and the biases $B^{(i)}_\alpha$. $\sigma(z)$ is the non-linear activation function, e.g. the ReLU $\sigma(z) = \max(z,0)$. As is commonly done, we use the "LeCun initialization" of the weights, for which they are scaled by a factor $1/\sqrt{h}$. It ensures that the limit of infinite width is not trivial, see e.g. discussions in \cite{jacot2018neural}. We represent the network  in Fig.~\ref{fig:architecture}.

Convolutional neural nets (CNNs), inspired from the primate brain, perform much better than FCs for a variety of tasks including image classification. Each hidden layer is composed of a number of channels,
each representing the data as a feature map -- itself an image.  CNNs perform the same operations at distinct locations in the image, and enforce that these operations are local. CNNs are thus more constrained than FCs, and they can be obtained from the latter by imposing that some weights are identical (to enforce translational invariance) and others zero (to enforce locality).


\paragraph{Learning Dynamics}
Weights are learnt by fitting the training set, which is done by minimising a loss or cost function ${\cal L}$. To simplify notations, we consider a binary classification problem, with a set of $P$ distinct training data denoted $\{(\mathbf{x}_\mu,y_\mu)\}_{\mu=1}^P$. The vector $\mathbf{x}_\mu$ is the input, which lives in a $d$-dimensional space, and $y_\mu=\pm 1$ is its label \footnote{It is straightforward to extend the discussion to multi-class classification problem, as well as to regression.}. ${\cal L}$ is a sum on all the training set of how each datum is well-fitted:
\begin{equation}
\label{loss}
    \mathcal{L}(\mathbf{\theta}) = \frac{1}{P} \sum_{\mu=1}^P \ell(y_\mu f(\mathbf{x}_\mu)),
\end{equation}
where $f$ is a model and $\ell$ is a decreasing function of its argument. Popular choices include the  linear ($\gamma=1$) or quadratic ($\gamma=2$) hinge loss $\ell(y_\mu f(\mathbf{x}_\mu))= \mathrm{max}\left(0, \Delta_\mu\right)^\gamma$ with $\Delta_\mu \equiv 1 - y_\mu f(\mathbf{x}_\mu)$, or the cross-entropy $\ell(y_\mu f(\mathbf{x}_\mu))= \log(1+\exp(-y_\mu f(\mathbf{x}_\mu)))$.  The hinge loss has the advantage that bringing the loss Eq.\ref{loss} to zero is equivalent to satisfying a set of constraints $\Delta_\mu<0$, corresponding to fitting the data by a margin unity. This fact is the basis for the analogy with other satisfiability problems encountered in physics, as discussed below.  This choice however does not influence significantly  performance for SOTA architectures on usual benchmarks ~\cite{Geiger18}.

Different procedures can then be used to minimise ${\cal L}$, starting from a random initialization of the weights (an important fact to keep in mind). In particular, gradient descent (GD) or stochastic gradient descent (SGD) are commonly used. For SGD,  only a (changing) subset of the terms entering the sum of Eq.\ref{loss} is considered at each minimisation time step. Various tricks can improve performance depending on the data considered, including early stopping (minimisation is stopped before the loss hits bottom) or weight decay (terms are then added to the loss to prevent weights to become too large).

\paragraph{Performance}
Once learning is done, generalization performance can be estimated from a test set (data not used to train, but whose distribution is identical to  the training set) by computing the probability to misclassify one new datum, the so-called test error $\epsilon$. How many data are needed to learn a task is characterized by the learning curve $\epsilon(P)\:$\footnote{A different network is trained for every $P$.}. Generally, it is observed that  the test error is well described by a power law decay $P^{-\beta}$ in the range of training set size $P$ available\footnote{For data sets where a finite fraction of the training set is mislabeled, one expects the test error to eventually plateau to a finite value. However, such a saturation of the test error is not seen in practice for various benchmarks, suggesting that such a noise is small in magnitude.} with an exponent  $\beta$ that depends jointly on  the data and the algorithm chosen. In~\cite{hestness2017deep}, $\beta$ is reported for SOTA architecture for several tasks: in \emph{neural-machine translation}  $\beta\approx0.3$--$0.36$; in \emph{language modeling} $\beta\approx0.06$--$0.09$; in \emph{speech recognition}  $\beta\approx0.3$ and in \emph{image classification} (ImageNet)  $\beta\approx0.3$--$0.5$.

\paragraph{ Deep nets learn a hierarchical  representation of the data}
Once learning took place, it is possible to analyse to which features in the data  neurons respond mostly to \cite{le2013building,zhou2014object}. Strikingly, findings are very similar to what is found in the visual cortex of primates:
neurons learn to respond to more and more abstract aspects of the data as the signal progresses through the network, as illustrated in Fig.\ref{fig:architecture}. This observation is believed to be an important aspect of why deep learning works,
yet understanding how abstract features are learnt dynamically, and how much it contributes to performance, remains a challenge.

\subsection{Why deep learning should not work}
\label{intro2}
\begin{figure}[htbp]
   \begin{center}
     \rotatebox{+0}{\resizebox{\textwidth}{!}{\includegraphics{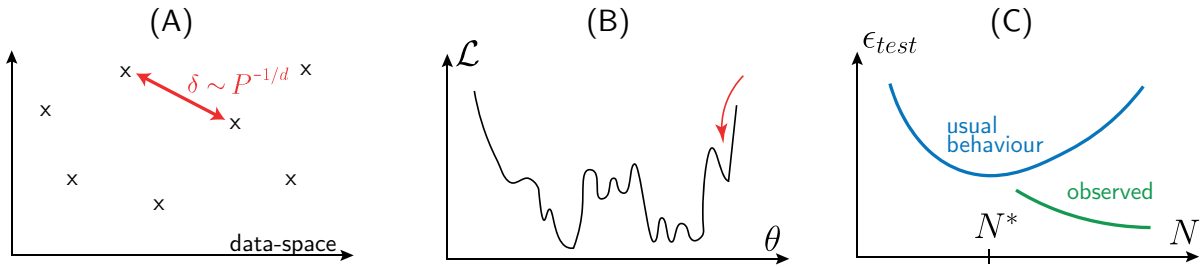}}}
     \caption{A: Curse of dimensionality. In high dimensions, every point lies far away from its neighbours, forbidding classification based on distances alone. B: Sketch of the loss landscape. Starting from a random initialization, the dynamics could get stuck in a bad minimum of the loss. C: Usual behaviour (blue line) of the test error  as a function of the number of parameters, expected to be minimal when $N$ is of order of the smallest value $N^*$ where data can be fitted. Green: observations indicate that for deep learning, increasing the number of parameters  generally improves behaviour, even in a regime where data are perfectly fitted.   }
     \label{curse}
   \end{center}
\end{figure}

\paragraph{Curse of dimensionality} 
General arguments suggest that $\beta$ should be extremely small --- and learning thus essentially impossible --- when the dimension $d$ of the data is large, which is generally the case in practice. For example in a regression task, if the only assumption on the target function is that it is Lipschitz continuous, then the test error cannot be guaranteed to decay faster than with an exponent $\beta \sim \nicefrac{1}{d}$ \cite{luxburg2004distance}. This \emph{curse of dimensionality} \cite{bach2017breaking} stems from the geometrical fact that the distance $\delta$ among nearest-neighbour data points decays extremely slowly in large $d$ as $\delta \sim P^{-\nicefrac1d}$, as depicted in the Fig.\ref{curse}.A. Thus, interpolation  or classification methods based on distances are expected to be very imprecise for generic data.

\paragraph{Bad minima in the loss landscape}
Another problem, which made deep learning less popular in the 90's, concerns  learning. The loss function has no guarantees to be convex, and the parameters space is high-dimensional. What prevents then  the learning dynamics to get stuck in poorly performing minima with high loss, as sketched in  Fig.\ref{curse}.B? In other words, under which conditions can one guarantee that training data are well fitted? Is the  loss landscape similar to the energy landscape of glassy systems in physics, where the number of minima is exponential in the number of degrees of freedom  \cite{reviewBB,Choromanska15}?

\paragraph{Over-parametrization}
Finally, neural nets are often trained in the {\it over-parametrized} regime, where the number of parameters $N$ is significantly larger than the number of data points $P$. In that regime, their capacity is very large: they can  fit the training set even if  labels are randomized \cite{zhang2016understanding}. Statistical learning theory then gives no guarantees that over-fitting does not occur, and that the model learnt has any predictive power. Indeed from statistics text books one expects a bell-shape learning curve relating the test error to the number of parameters of the model, as depicted in  Fig.\ref{curse}.C.  A very puzzling aspect of deep learning is that increasing $N$ passed the point where all data are perfectly fitted does not destroy predictive power, but instead  {\it improves} it ~\cite{neyshabur2017geometry,neyshabur2018towards,bansal2018minnorm,advani2017high} as shown in Fig.\ref{curse}.

\subsection{Current insights on these paradoxes}
\label{intro3}

\paragraph{No bad minima in over-parametrized networks}
Recently, it was realised that the landscape of deep learning is not glassy after all, if the number of parameters is sufficient. Theoretical works that focus on nets with a huge number of parameters (and one hidden layer), and empirical work with more realistic architectures ~\cite{Freeman16,Hoffer17,Soudry2016,Cooper18}, support that the loss function is characterized by a connected level set: any two points in parameter space with identical loss are connected by a path in which the loss is constant. Moreover, empirical studies of the curvature of the loss landscape (captured by the spectrum of the Hessian) ~\cite{Sagun16,sagun2017empirical,Ballard17} and of SGD dynamics~\cite{Lipton16,Baity18} reveal that the landscape displays a large number of flat directions, even at its bottom. It is not the case for under-parametrized networks \cite{Baity18}. 

These observations raise key questions. {\it Is there a phase transition in the landscape geometry as the number of parameters grows? If so, what is its universality class? How does it affect performance?}

\paragraph{Implicit regularization and infinite width nets}

\begin{figure}[htbp]
   \begin{center}
     \rotatebox{+0}{\resizebox{13.5cm}{!}{\includegraphics{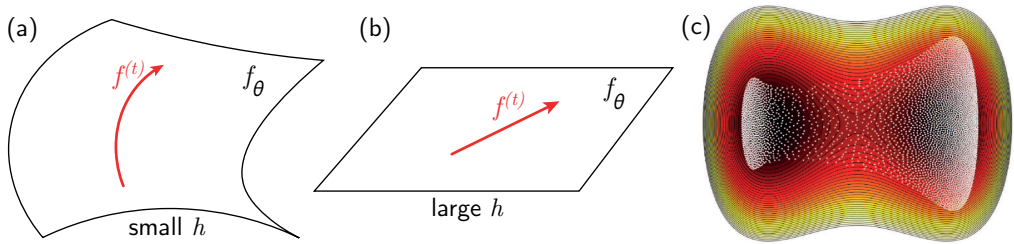}}}
     \caption{A: sketch of the dynamics in function space for a finite width net. As weights change the function evolves along the black manifold of expressible functions, which is curved. B: in the infinite width limit, this manifold becomes flat  even though the function  evolves by ${\cal O}(1)$: the relationship between function and weights became linear. C: interacting particles in a potential. When their number is large, describing their density instead of their individual motion becomes favorable (image graciously shared by Eric Vanden-Eijnden).}
     \label{ntk}
   \end{center}
\end{figure}

How can over-parametrized networks be predictive, if they can fit any data? It must imply that the GD or SGD dynamics lead to specific minima of the loss landscape where the function is more regular than for generic minima-- a phenomenon coined {\it implicit regularization} \cite{neyshabur2017geometry,neyshabur2018towards}.
As sketched in Fig.\ref{curse}, performance is best as $N\rightarrow\infty$, triggering a huge interest in that limit. To which algorithms does GD  correspond to in that case?


{\bf NTK:} The propagation of the input signal through infinite-width FC nets  at initialization is now well-understood. If the weights -- defined in Eqs.\ref{eq:recnnet} --  are initialized as i.i.d. random variables with zero mean and variance one (a set-up we consider throughout this work), the output function $\tilde f(x)$ is a Gaussian random process. Its covariance  can be computed recursively~\cite{Neal1996,williams1997computing,Lee2017,matthews2018gaussian,novak2018bayesian,yang2019scaling}.

Very recently it was realised  that the {\it learning dynamics} also simplifies in this limit ~\cite{jacot2018neural,Du2019,Allen-Zhu2018,lee2019wide,arora2019exact,park2019effect}.
The key insight of \cite{jacot2018neural} is that  the output  becomes a linear function of the weights then, as sketched in Fig.\ref{ntk} A,B.  Physically, very tiny changes of weights can interfere positively
and change the output by ${\cal O}(1)$ which is sufficient to learn, but is not sufficient to change the gradient $\nabla_\theta \tilde f$. 

More formally, the gradient flow dynamics at finite time can always be described by the \emph{neural tangent kernel} (NTK) defined as:
\begin{equation}
\Theta(\theta,x,y) = \nabla_\theta \tilde f_\theta(x) \cdot \nabla_\theta \tilde f_\theta(y),
\label{eq:ntkformula}
\end{equation}
where $x,y$ are two inputs and  $\nabla_\theta$ is the gradient with respect to the parameters $\theta$. Specifically, $\partial \tilde f(x)/\partial t$ can be expressed as a linear combination of the  $\Theta(\theta,x,x_i)$; i.e. $\tilde f$ evolves within a space of dimension $P$. In general,  $\Theta$ depends on  $\theta$ and thus  on time and on the randomness of the initialization.  Yet as $h\to\infty$, $\Theta(\theta,x,y)$ converges to a well-defined limit independent of initialization, and does not vary in time: deep learning in that limit is thus essentially a kernel method \cite{jacot2018neural}.

{ \bf Feature learning (hydrodynamic) regime:}
The infinite width limit can be taken differently by adding a factor $1/\sqrt{h}$ in the definition of the output. Then the weights must change significantly for the output to become ${\cal O}(1)$ and fit the data. This recently discovered limit is called ``mean field'', ``rich" but also ``feature learning regime" in the literature, because the neurons learn how to respond to different aspects of the input data, as in Fig.\ref{fig:architecture} -- whereas, in the NTK limit, the neuron's response evolves infinitesimally. This regime has been studied in several works focusing mostly on one-hidden layer networks~\cite{mei2018mean,rotskoff2018neural,chizat2018,sirignano2018mean,mei2019mean,nguyen2019mean,sirignano2020mean}, with recent development for deeper nets, see e.g.  \cite{nguyen2020rigorous}. In this setting the output function for a one hidden layer reads:
\begin{equation}
\label{111}
\tilde f_\theta(x) = \frac1h \sum_{i=1}^h W^{(2)}_i \sigma(\frac{1}{\sqrt{d}} W^{(1)}_i\cdot x + B_i),
\end{equation}
The law of large number can then be invoked to replace Eq.\ref{111} by an integral:
\be
\label{112}
\tilde f_\theta(x) \to \int\mathrm{d}W^{(2)} \mathrm{d}W^{(1)} \mathrm{d}B\, \rho(W^{(2)}, W^{(1)}, B)\; W^{(2)} \sigma(\frac{1}{\sqrt{d}} W^{(1)}\cdot x + B)
\ee
 where  $\rho$ is the density of parameters. It is then straightforward to show that gradient flow leads to a dynamics on $\rho$ of the usual type for conserved quantities: it is the divergence of a flux $\partial \rho/\partial t=-\nabla \cdot J$. Here the divergence is computed with respect to the set of  weights associated with a single neuron, and the flux follows $J=\rho \Psi(W^{(2)},W^{(1)},B; \rho_t)$ where $\Psi$ is some function that can be expressed in terms of the loss ~\cite{mei2018mean}. This formulation is equivalent to the hydrodynamics description of interacting particles in some external  potential. Fig.\ref{ntk}.C illustrates that when the number of particles (or neurons) is large, $\rho$ is a more appropriate description than keeping track of all the particle positions (the weights in that case).

{\bf Lazy training:} The fact that deep learning converges to well-defined algorithms as $h$
diverges, explains why performance converges to a well-defined value in that limit (which will depend on the limit considered), as sketched in Fig.\ref{curse}.  Yet, this distinction between a regime where the NTK does not change, and one where it evolves and features are learnt can be made at finite $h$. Chizat and Bach proposed a model of the form \cite{chizat2019lazy}:
\begin{equation}
f_\theta(x) \equiv \alpha\left[\tilde f_\theta(x) - \tilde f_{\theta(t=0)}(x)\right],
\label{eq:bachmodel}
\end{equation}
$\alpha={\cal O}(1)$ corresponds to the NTK initialization  and $\alpha={\cal O}(h^{-\nicefrac12})$ to the mean-field limit. 
In the over-parametrised regime (our work below defines it sharply) where data are fitted, infinitesimal changes of weights are sufficient to learn when
 $\alpha\rightarrow\infty$ at finite $h$, and the NTK does not evolve in time (but has fluctuations at initialization).
 This regime is sometimes referred to as lazy training, in our context we will interchangeably denote it the NTK regime.


Overall, these findings are recent breakthrough in our understanding of neural nets. Yet they ask many questions that are central to this review.  {\it If two limits exist, which one best characterises  neural networks that are used in practice?  Which limit leads to a better performance? How does it depend on the architecture and data structure? Which effect causes the improvement of performance as $h$ increases?}

\paragraph{Curse and invariance toward diffeomorphisms}
    Mallat and Bruna \cite{bruna2013invariant, mallat2016understanding} proposed that  invariance toward smooth deformations of the image, i.e. diffeomorphisms,  may allow deep nets to beat the curse of dimensionality. 
    Specifically, consider the case where the input vector $x$ is an image. It can be thought as a function $x(s)$ describing intensity in position $s$, where $s\in[0,1]^2 $ \footnote{This space is in fact 
    discrete due to the finite number of pixels, but this does not alter the discussion.}. A diffeomorphism  is a  bijective deformation that changes the location $s$ of the pixels to $s'=
    \tau(s)$.  In our review below, it will be useful to introduce the pixel displacement field $\xi(s)=\tau(s)-s$, analogous to the displacement field central to elasticity. 
    We denote by $T_\tau[x]$ the image deformed by $\tau$, i.e. $T_\tau[x](s)=x(\tau^{-1}(s))$. One expects that smooth diffeomorphisms do not affect the label of an image:
    $Prob(y(x)\neq y(T_\tau[x]))\ll1$ if $\|\nabla_s \xi\|\ll1$.
    
    Mallat and Bruna could handcraft CNNs, the "scattering transform",  that are insensitive to smooth diffeomorphisms and perform well:
   \be
    \label{diffeo}
   | f(x)-f(T_\tau[x]) | \leq C_0 \|\nabla_s \xi\|.
   \ee
 They hypothesised that during training, CNNs learn to satisfy Eq.\ref{diffeo}.  In this view, by becoming insensitive to many aspects of the data irrelevant for the task, CNNs effectively reduce the data dimension and make the problem tractable.     


A  limitation of this framework is that it says little about how invariants are learnt dynamically. Yet, it is the center of the problem. FC nets are more expressive than CNNs,
and nothing a priori prevents them from learning these invariants -- yet, they presumably don't,  since their performance do not compare with CNNs. 
Along this route, an interesting question is how to
develop observables to characterise empirically the dynamical emergence of invariants. Attempts have been made using mutual information approaches \cite{shwartz2017opening} which display problems in such a
deterministic setting \cite{saxe2019information}; or measures based on the effective dimension of the neural representation of the data \cite{ansuini2019intrinsic,recanatesi2019dimensionality} which are informative but can sometimes lead to counter-intuitive results \footnote{The effective dimension is defined from how the distance $\delta$ among neighboring points varies with their number $P$, i.e. $\delta\sim P^{-1/d_\text{eff}}$. Its interpretation can be delicate, as it is observed to sometimes  increase as the information propagates in the network, which cannot be the case  for the true dimension of the manifold representing the data.}. In Section \ref{S5} we discuss other observables based on kernel PCA of the NTK. 

\begin{figure}[!ht]
   \begin{center}
     \scalebox{1}{ \import{figures/}{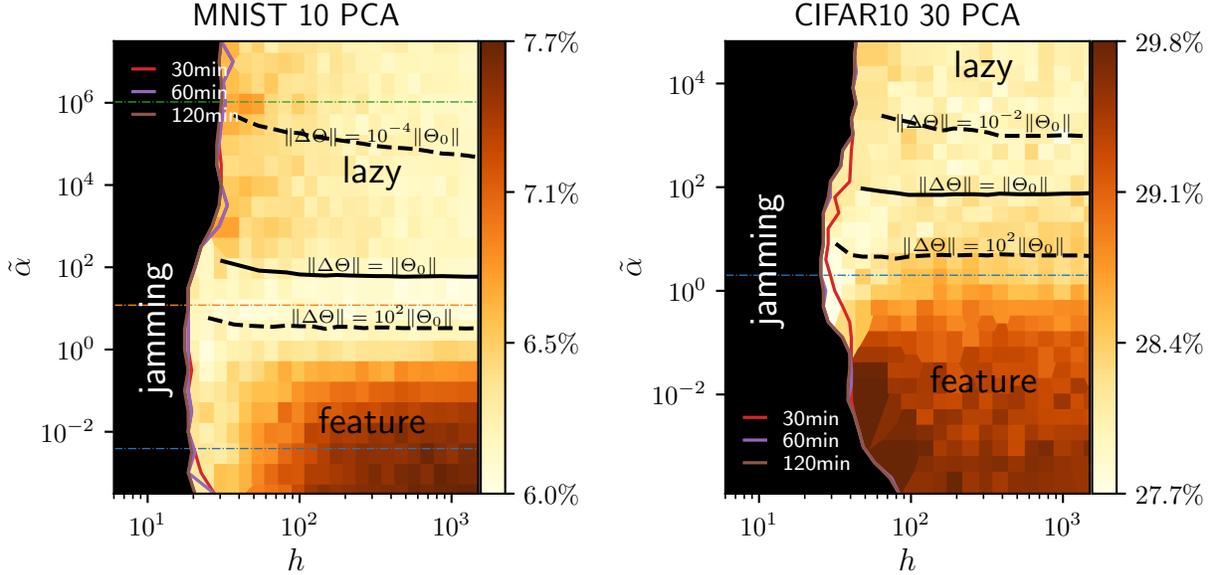} }
     \caption{Phase diagram for the performance of deep learning in the width $h$ {\it v.s.} rescaled initialization magnitude $\tilde{\alpha}=\sqrt{h}\alpha$ plane, where $\alpha$ is defined in Eq.\ref{eq:bachmodel}.  The architecture corresponds to a fully-connected neural network with two hidden layers trained with gradient descent  \cite{geiger2019disentangling}. Binary classification was performed for \underline{Left} MNIST binary (odd/even)  and \underline{Right} CIFAR10 binary (classes have been regrouped in two sets). The training sets have been reduced to 5000 samples. These data sets were first projected on 10 and 30 PCAs, respectively.  This dimensionality reduction of the problem is useful to study the jamming transition\protect\footnotemark 
        but not necessary to investigate the over-parametrized regime \cite{geiger2019disentangling}. This choice does not affect qualitatively the geometry of the diagram however. The colormap shows the generalization of the ensemble average over 15 samples. 
     The full black line indicates the cross-over from feature to lazy training. It is defined as the location for which the change of the NTK after training $\Delta \Theta=\Theta(t=\infty)-\Theta(0)$ is equal in norm to its magnitude at initialization, i.e.  $\|\Delta \Theta\|/\|\Theta(0)\|=1$.
     Choosing other values for this ratio (see dashed lines) does not affect qualitatively the geometry of the cross-over line.
     The vertical color lines indicate the jamming transition where the training loss hits zero (in the black region the training loss does not reach zero). Its precise position depends on the time the simulation is allowed to run \cite{Geiger18}, as indicated in legend (using a GPU V100 and the gradient flow algorithm of \cite{geiger2019disentangling} (gradient descent with adaptive learning rate)).
     It is expected and corresponds to the usual critical slowing down occurring near phase transitions.
     Horizontal dashed lines correspond to the measures done in Fig.\ref{overfitting}.
     The code used to generate these data is available at \url{https://doi.org/10.5281/zenodo.4322828}. }
     \label{phase_space}
   \end{center}
\end{figure}

\subsection{Organization of the manuscript}
\label{intro4}

To think about the results above and the questions they raise in a unified manner, it is useful to consider Fig.\ref{phase_space}. It shows  novel empirical data for  the performance of neural nets in the plane $(h,\tilde{\alpha})$ where $h$ is the width and $\tilde{\alpha}$ the rescaled initialization amplitude $\tilde{\alpha}=\sqrt{h}\alpha$. Specifically, the ensemble test error of  fully-connected nets learning MNIST (a data set of images of digits \cite{lecun-mnist}) and CIFAR10 (images of planes, dogs, etc...) respectively with gradient descent is shown. This ensemble test error is computed by preparing, for each point in the phase diagram, $M=15$ nets with different initialization. The test error of the mean predictor $\langle f\rangle$ over the 15 trained nets is then shown. The black region corresponds to the under-parametrized regime where the  loss of individual nets does not converge to zero after learning. As discussed below, its boundary corresponds to a line of critical points corresponding to a "jamming transition" where the system stops displaying a rough landscape.
In the colored over-parametrized regime, the black line indicates a cross-over separating the lazy training regime (where the total change of the NTK of individual nets during learning is small in relative terms) from a feature learning regime (where the NTK evolves significantly). 

\begin{figure}[!ht]
   \begin{center}
     \scalebox{1}{ \import{figures/}{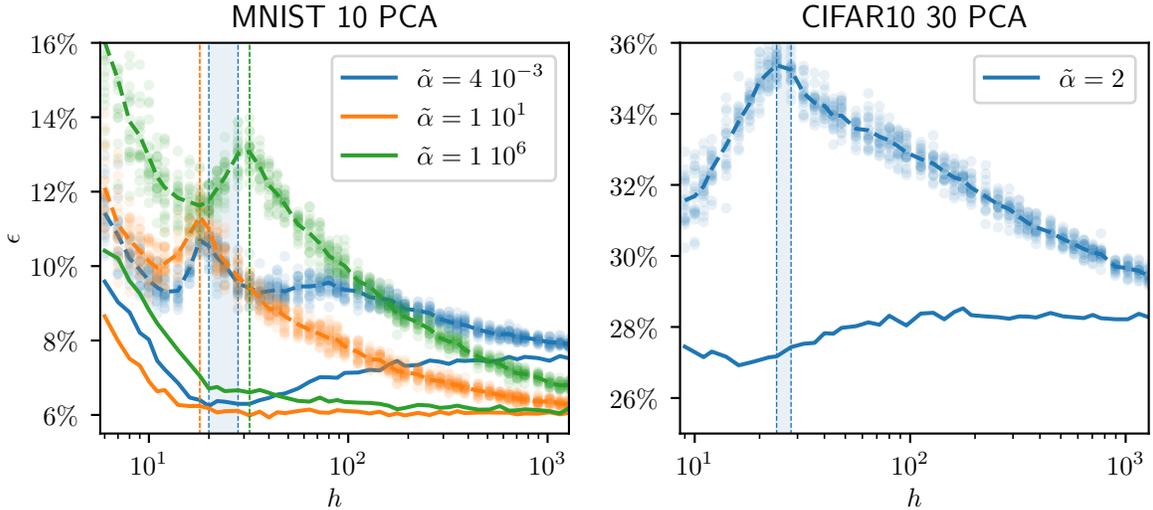} }
     \caption{Comparison of the generalization error and the ensemble generalization error for the same data, architecture and learning dynamics used in Fig.\ref{phase_space}. Values of $\tilde \alpha$ used are indicated in legends, and corresponds to the colored horizontal dashed lines in Fig.\ref{phase_space}. The ensemble average is done over 20 samples. The vertical lines indicate the location, or interval of locations, where the jamming transition occurs. Note that  ensemble averaging improves a lot performance at intermediate width, and eliminates the double descent both in the lazy or feature learning regime.}
     \label{overfitting}
   \end{center}
\end{figure}

Fig.\ref{overfitting} shows the curves for the test error and the ensemble test error for three values of $\tilde \alpha$, as $h$ varies. At the jamming transition, the test error displays a peak as observed in \cite{Spigler18}.
A similar peak occurs in regression problems \cite{advani2017high}, where it has a simpler origin and occurs when the number of parameters matches the number of data. This shape for the test error has since then been coined double-descent \cite{belkin2019reconciling}. Interestingly, as shown in Fig.\ref{overfitting} this phenomenon occurs independently of the value of $\tilde \alpha$ (and thus of the over-parametrized regime one enters into passed jamming) and vanishes when the ensemble average is taken, as observed in \cite{geiger2019scaling} and recently confirmed in \cite{lee2020finite}.
\footnotetext{Otherwise, for FC architectures the value of $h$ at which the jamming transition occurs is a few neurons, leading to new effects.}

Our goal is to review, in non-technical terms, concepts and arguments justifying such a phase diagram and discuss the learning of invariants in this context.
In section \ref{S2}, we will argue that the boundary of the black region  in Fig.\ref{phase_space}  is a line of critical point, analogous to the jamming transition occurring in  repulsive particles with finite-range interactions. For wider nets, the landscape is not glassy and displays many flat directions. We will provide a simple geometric argument justifying why this transition in deep nets fall into the universality classes of ellipsoid (rather than spherical) particles, which fixes the properties of the landscape (such as the spectrum of the Hessian) near the transition. We will mention an argument \`a la Landau justifying the cusp in the test error at that point. In Section \ref{S3}, we will provide a quantitative explanation as to why the test error keeps improving as the width increases passed this transition.  Increasing width turns out to eliminate the noise due to the random initialization of the weights,  eventually  leading  to the well-defined algorithms introduced above.  Ensemble averaging at finite width efficiently eliminates this source of noise as well as the double descent, as shown in Fig.\ref{overfitting}.  In Section \ref{S4}, we will explain why the cross-over between the lazy and feature learning regimes corresponds asymptotically to a flat line in Fig.\ref{phase_space}. We will review observations that  lazy training   outperforms feature learning for standard data sets of images   for fully connected architectures, but not for CNNs architectures, corresponding to a larger learning curve exponent $\beta$. In Section \ref{S5}, we review arguably the simplest model in which a neural net learns invariants in the data structure, by considering that labels do not depend on some directions in input space.  As observed in real data, two distinct training exponents $\beta$ in the lazy and feature learning can then be computed. We conclude by discussing open questions. 

\section{ Loss landscape and Jamming transition}
\label{S2}
\begin{figure*}
    \centering
    \setlength{\unitlength}{0.1\textwidth}
    \begin{picture}(10,4)
    \put(0,2){\def\svgwidth{0.24\textwidth}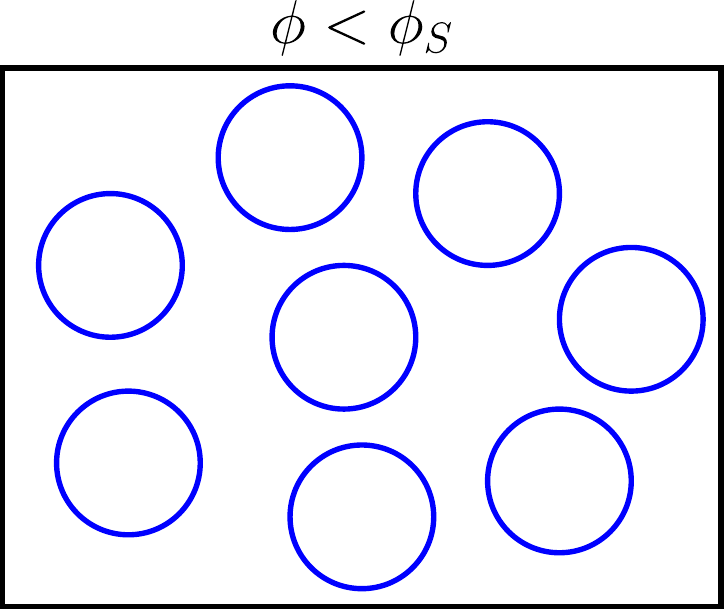}\put(2,3.9){(a)}
    \put(2.5,2){\def\svgwidth{0.24\textwidth}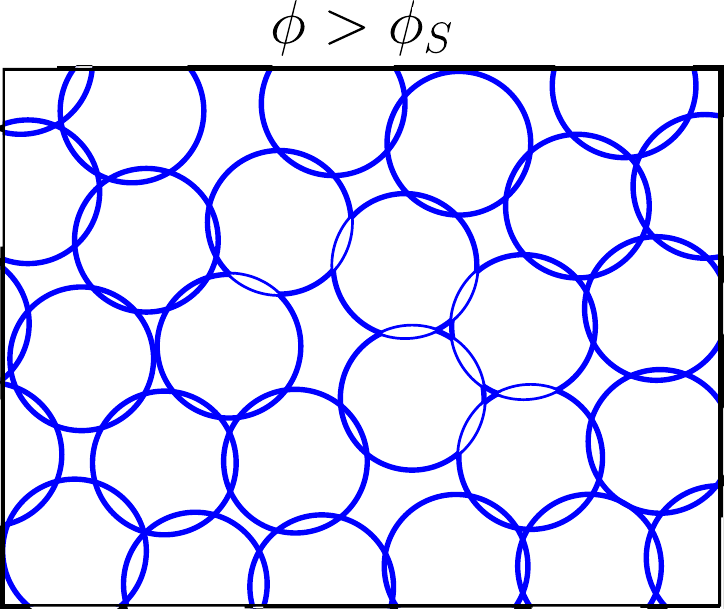}\put(4.5,3.9){(b)}
    \put(5,2){\def\svgwidth{0.24\textwidth}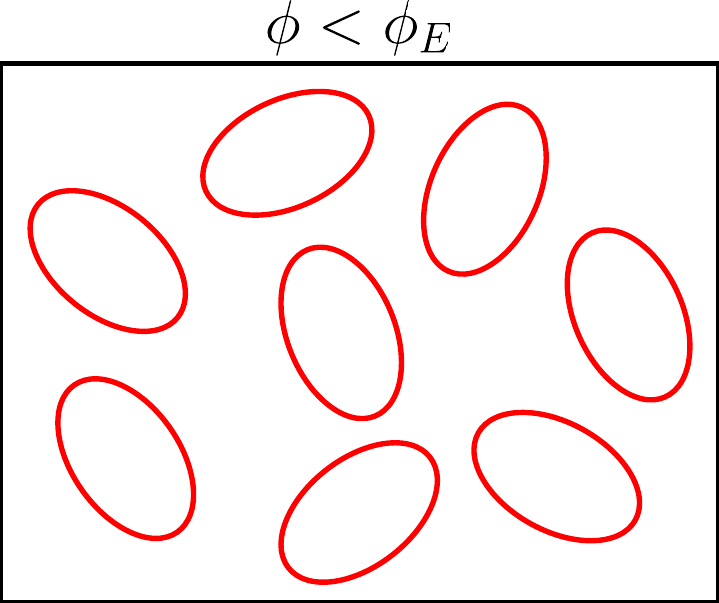}\put(7,3.9){(c)}
    \put(7.5,2){\def\svgwidth{0.24\textwidth}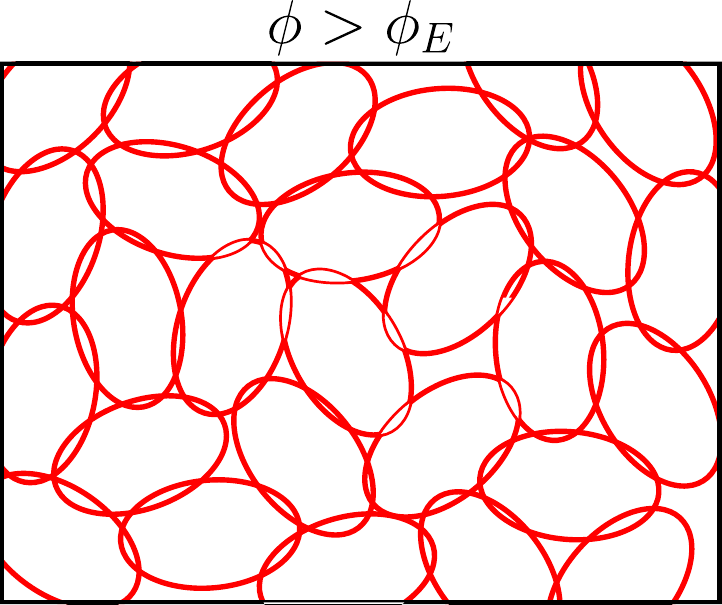}\put(9.5,3.9){(d)}
    \put(0,-0.05){\def\svgwidth{0.24\textwidth}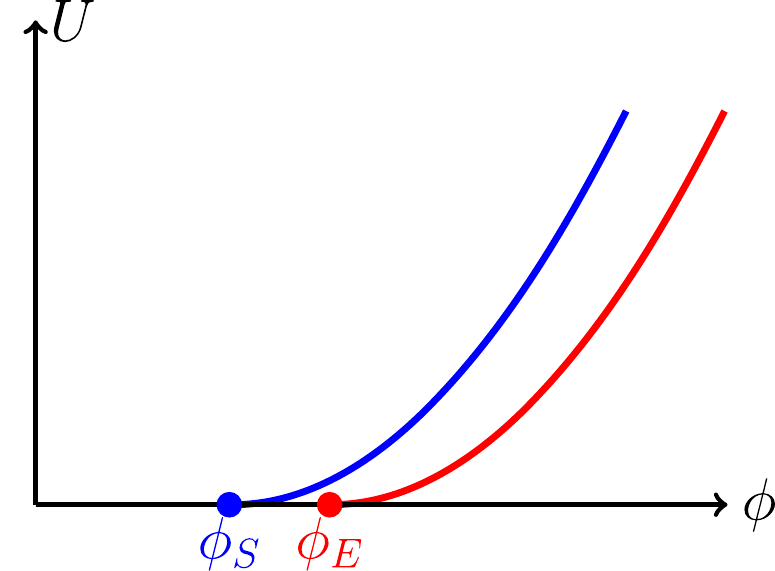}\put(2,1.6){(e)}
    \put(2.5,0.1){\def\svgwidth{0.24\textwidth}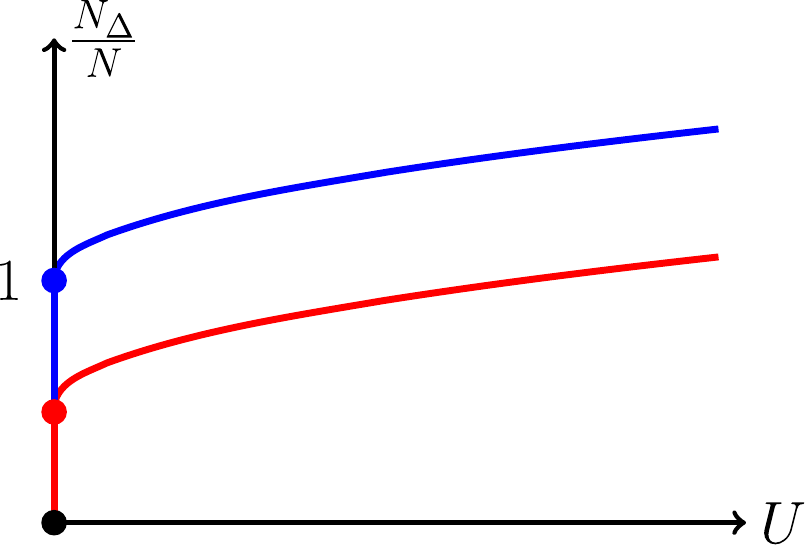}\put(4.5,1.6){(f)}
    \put(5,0){\def\svgwidth{0.24\textwidth}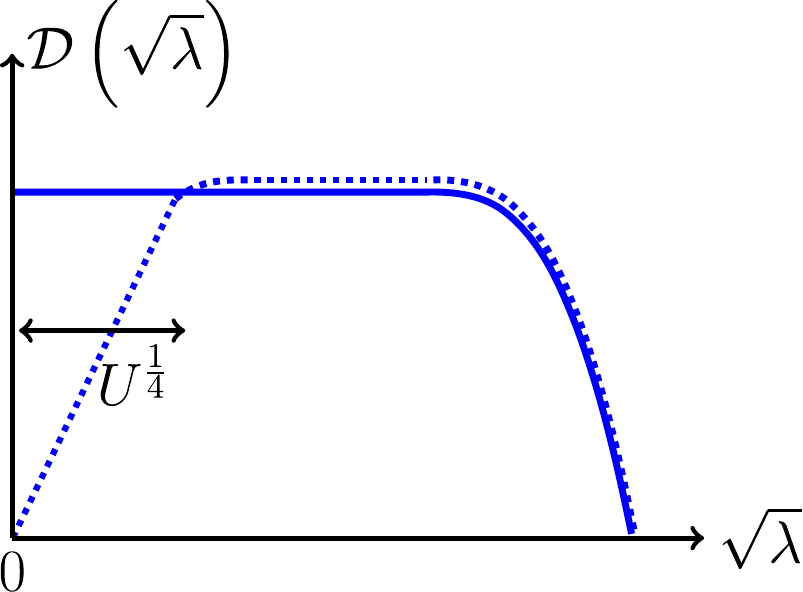}\put(7,1.6){(g)}
    \put(7.5,0){\def\svgwidth{0.24\textwidth}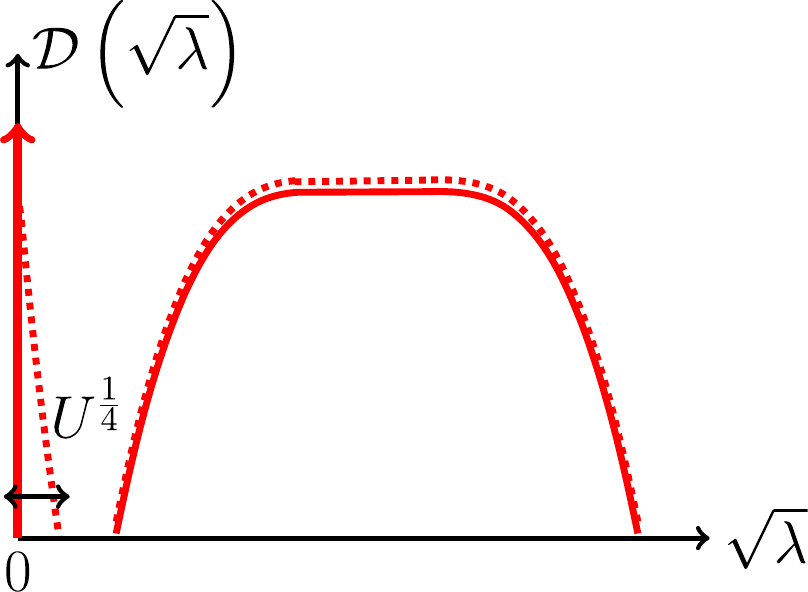}\put(9.5,1.6){(h)}
    \end{picture}
    \caption{Sketch of the jamming transition for repulsive spheres and ellipses. (a,b,c,d) Both systems transition from a fluid to a solid as the density passes some threshold, noted $\phi_S$ for spheres and $\phi_E$ for ellipses. (e) For denser packings, the potential energy ${\cal U}$ becomes finite. (f) The ratio $N_\Delta/N$ between the number of  particles in contact $N_\Delta$ (corresponding to unsatisfied constraints) and the number of degrees of freedom $N$ jumps discontinuously to a finite value, which is unity for spheres but smaller for ellipses. (g,h) This difference has dramatic consequence on the energy landscape, in particular on the spectrum of the Hessian. In both cases, the spectrum becomes non-zero at jamming, but it displays a delta function with finite weight for ellipses (indicating strictly flat directions), followed by a gap with no eigenvalues, followed by a continuous spectrum (h, full line). For spheres, there is no delta function nor gap (g, full line). As one enters the jammed phase, in both cases a characteristic scale $\lambda\sim \sqrt{{\cal U}}$ appears in the spectrum (g and h, dotted lines). From \cite{Geiger18}.   \label{fig1}}
\end{figure*}

Minimizing a loss is very similar to minimizing an energy.   Describing the energy landscape of physical systems is a much studied problem, especially in the context of glasses. Progress was made on this topic in the last fifteen years by considering  finite range interactions, for which bringing the energy to zero is equivalent to  satisfying a set of constraints.
In that case, the landscape is controlled by a critical point \cite{Ohern03,Wyart05b,Liu10}, the so-called jamming transition.  It occurs as the  particle density $\phi$ increases. For $\phi>\phi_c$, a gradient descent from a random initial positions of the particles gets stuck in one -- out of many -- meta-stable states, corresponding to a glassy solid as depicted in Fig.\ref{fig1}B,D. For lower densities, the gradient descent reaches zero energy, as sketched in Fig.\ref{fig1}.E : particles can freely move without restoring forces, and the landscape has many flat directions. It corresponds to the situations depicted in  Fig.\ref{fig1}A,C. 

\begin{figure}[!ht]
   \begin{center}
     \scalebox{0.8}{ \import{figures/}{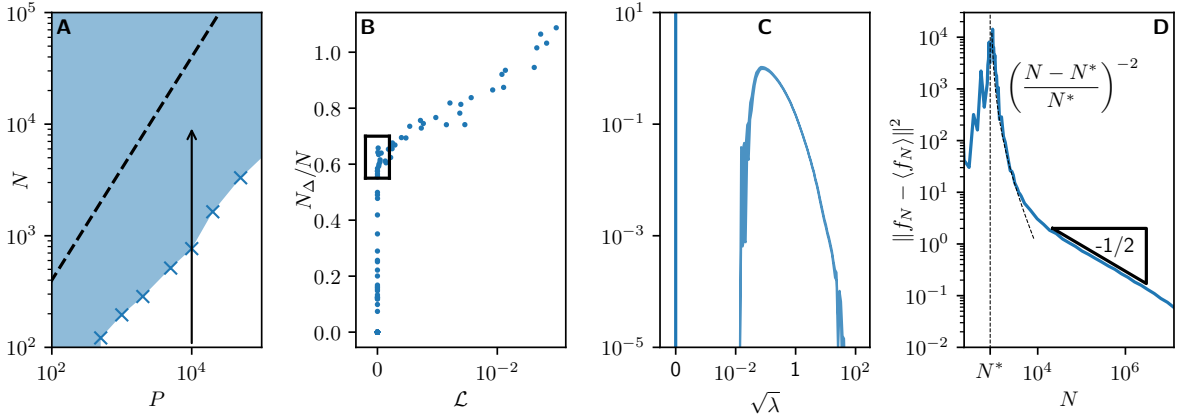} }
     \caption{\textbf{A} Phase diagram indicating where the hinge loss hits zero (in blue) or not (in white) at the end of training, as the number of parameters $N$ and  training points  $P$ vary, for the MNIST data set discussed above (binary classification, 10 PCAs) trained with a FC nets with 3 hidden layers. 
     \textbf{B}  Number  of unsatisfied patterns $N_\Delta/N$ as a function of the loss at the end of training, following the arrow indicated in panel A. Here  $N$ is fixed, and $P=1, \dots 70k$. 
     \textbf{C} The spectrum of the dominant term in the Hessian near jamming $H_0$ (averaged over the run at jamming, in the rectangle of \textbf{B}) presents a delta function in zero and a gaped continuous spectrum at high frequencies, a generic feature of hypostatic jamming.
     \textbf{D} Fluctuations of the output function along the arrow shown in \textbf{A}, averaged over 20 initial conditions. It decays asymptotically as a power-law $N^{-1/4}$ for the MNIST data set. At jamming, these fluctuations explode, consistent with a quadratic singularity. From \cite{Geiger18,geiger2019scaling}.}
     \label{doubledescent}
   \end{center}
\end{figure}

 There are two universality classes for jamming, leading to distinct properties for the curvature of the landscape (i.e. the spectrum of the Hessian) \cite{Wyart05a,DeGiuli14,Franz15,Brito18a,Mailman09,Zeravcic09}, for the structure of the packing obtained \cite{wyart2012marginal,Lerner13a,Charbonneau14,Charbonneau14a} and for the dynamical response to a perturbation \cite{lerner2012unified,degiuli2015unified}. Spheres and ellipses fall in distinct classes as illustrated in Fig.\ref{fig1}. More generally, the jamming transition occurs  generically in satisfiability problems with continuous degrees of freedom (it can be defined with discrete degrees of freedom \cite{Krzakala07}, but then differs qualitatively; in particular, the present discussion does not apply to the discrete case). It occurs in the perceptron \cite{Franz16,Franz17, franz2018jamming} but also for deep nets \cite{Geiger18,Spigler18}. 
 
We will recall below a geometric argument introduced in \cite{Geiger18} determining the universality class of the jamming transition. For deep nets, jamming  belongs to the universality class of ellipses \cite{Geiger18,Spigler18}. The spectrum of the loss near the jamming transition  displays zero modes, a gap and a continuous part, as measured in Fig\ref{doubledescent}.C.  Another  implication of this analogy is the number $N_\Delta$ of data whose margin is smaller than unity after learning,  which contribute to the loss. These data are conceptually similar to the support vectors central to SVM algorithms. For particles, $N_\Delta$ corresponds the number of pairs of particles still overlapping after energy minimization. As illustrated in Fig\ref{doubledescent}.B, $N_\Delta/N$ jumps from zero to a value strictly smaller than one at jamming where the loss becomes positive, precisely as ellipses do. 

\paragraph{Geometric argument fixing the universality class:}
Here we seek to give a simple intuition of the result of \cite{Geiger18} (a more rigorous argument can be found there). We consider continuous satisfiability problems where one seeks to minimize an energy or loss function of the type:
\be
\label{1}
{\cal U}=\sum_{\mu \in m} \frac{1}{2}  \Delta_\mu^2.
\ee
where the sum is made on all constraints that are not satisfied, corresponding to $\Delta_\mu>0$. The quantities $\Delta_\mu$ can depend in general on the $N$ degrees of freedom of the system. For systems of particles, ${\cal U}$ is an energy and $\Delta_\mu$ is the overlap between a pair $\mu=(i,j)$ of particles $i$ and $j$. For  spheres it reads $\Delta_{ij}=2R-r_{ij}$ where $R$ is the particle radius and $r_{ij}=|\!|{\bf r}_i - {\bf r}_j|\!|$ the distance between particles $i$ and $j$. In that case, $N$ is   the number of particles times the spatial dimension. For the perceptron or deep nets, ${\cal U}$ corresponds to a quadratic hinge loss generally denoted ${\cal L}$ defined in Eq.\ref{loss} (it is straightforward to extend these arguments to other hinge losses with $\gamma>1$ \cite{Geiger18,Ohern03}, the case $\gamma=1$ of the linear hinge shares many similarities with those but also display interesting differences \cite{franz2019critical}). As defined in the introduction, in that case $\Delta_\mu=1-f({\bf x}_\mu)y_\mu$.



As the jamming transition is a singular point, it is useful -- but not strictly necessary \footnote{Particles can always sit right at jamming if an infinitesimal pressure is applied to the system. In the context of neural nets, it can also be achieved by introducing an additional term in the loss of the type $\lambda ||\theta(t)-\theta(t=0)||^2$ penalizing the square evolution of the weights (called "weight decay")  of vanishing magnitude $\lambda\rightarrow 0^+$.} -- to think of  approaching from the glassy (large density $\phi$ or under-parametrized) phase. It corresponds to ${\cal U}\rightarrow 0$ as sketched in Fig.~\ref{fig1} E,F,  implying that $\Delta_\mu  \rightarrow 0$ $ \forall \mu \in m$. As argued in
\cite{Tkachenko99}, for each $\mu \in m$ the constraint $\Delta_\mu  = 0$ defines a manifold of dimension $N-1$. Satisfying $N_\Delta$ such equations thus generically leads to a manifold of solutions of dimension $N-N_\Delta$. Imposing that  solutions exist thus implies that, at jamming, one has:
\be
\label{2}
N_\Delta \leq N\,.
\ee
Note that this argument implicitly assumes that  the $N_\Delta$ constraints are independent, see \cite{Geiger18} for more discussions.

An opposite bound can  be obtained  by considerations of stability, by imposing that in a stable minimum the Hessian must be positive definite~\cite{Wyart05a}.
The Hessian is an $N\times N$ matrix which can be written as:
\be
\label{3}
{\cal H}_U=\sum_{\mu \in m} \nabla\Delta_\mu \otimes \nabla\Delta_\mu + \sum_{\mu \in m} \Delta_\mu {\cal H}_{ \Delta_\mu} \equiv {\cal H}_0 + {\cal H}_p\,,
\ee
where ${\cal H}_{ \Delta_\mu}$ is the Hessian of $\Delta_\mu$, and ${\cal H}_0$ and ${\cal H}_p$ correspond to the first and second sum, respectively. ${\cal H}_0$ is positive semi-definite, since it is the sum of $N_\Delta$ positive semi-definite matrices of rank unity; thus ${\rm rk}({\cal H}_0) \leq N_\Delta$, implying that the null-space of ${\cal H}_0$ is at least of dimension $N-N_\Delta$. ${\cal H}_p$ becomes very small approaching jamming, since the $\Delta_\mu$'s vanish. Let us denote by $N^-$ the number of negative eigenvalues of ${\cal H}_p$. Requiring that ${\cal H}_U$ has no negative eigenvalues thus implies:
\be
\label{2bis}
N_\Delta \geq N^-\,.
\ee
For \emph{spheres} \cite{Wyart05b} (as well as for the perceptron if a negative margin is used while constraining the norm of the weight vector), ${\cal H}_p$ is negative definite and $N^-=N$. This results stems from the fact that ${\cal H}_{ \Delta_\mu}$ is negative semi-definite. Indeed ${\cal H}_{ \Delta_\mu}$ characterizes the  second order change of overlap between particles, and is negative because the distance between spheres moving transversely to their relative direction  always increase following Pythagoras theorem. In that case we thus have $N_\Delta \geq N$. Together with Eq.~(\ref{2}) that leads to  $N_\Delta = N$: as spheres jam  the number of degrees of freedom and the number of constraints (stemming from contacts) are equal, as empirically observed~\cite{Ohern03}.  This property is often called {\it isostaticity}. 

However, in other problems such as ellipses and deep nets, ${\cal H}_{ \Delta_\mu}$ and thus ${\cal H}_p$ have positive eigenvalues. Indeed the overlap between two ellipses can {\it increase} if one of them rotates. Likewise, for a fully-connected Relu net and random data at initialization, ${\cal H}_p$ has a symmetric spectrum and $N^-\approx N/2$. Generically one expects then jamming to occur with $N_\Delta<N$ as sketched for ellipses in Fig.\ref{fig1} and shown empirically for neural nets learning MNIST in  Fig.\ref{doubledescent}. Jamming is then referred to as  "hypostatic". The associated consequences on the spectrum of the hessian shown on the same figures are derived in specific cases in \cite{wyart2010scaling, Brito18a,Franz15}: it always presents a delta function in zero and a gap at jamming.

\paragraph{Effect of the number of training data $P$}
The location of the jamming transition $N^*(P)$ depends on $P$. This dependence is linear for random data but sub-linear for structured data \cite{Geiger18}, as exemplified in Fig.\ref{doubledescent}.A. Denoting $N^-=C_0 N$, and using Eq.\ref{2bis} together with $N_\Delta\leq P$, we obtain $N^*(P)\leq P/C_0$, guaranteeing convergence to a zero loss for a number of parameters linear in $P$ if $C_0>0$.  We  conjecture that $C_0$ to remain bounded by a strictly positive value for large $N$ for  generic data and architectures used in practice \footnote{This fact does not hold for linear or polynomial activation function, see discussion in \cite{Geiger18}.}.  $C_0$ can be measured a posteriori, leading to the bound corresponds to the dotted line in Fig.\ref{doubledescent}.A. $C_0$ a priori depends on the choice of architecture,  dynamics and data set. Controlling its value  a priori is yet out of reach \footnote{Controlling $H_0$ and $H_p$ during learning was done only in the limit $N\rightarrow \infty$ which does not apply near jamming \cite{jacot2019hessian}.}, and would lead to a rather tight guarantee of convergence toward a global minimum of the loss.

\paragraph{Effect of the scale of initialization {\bf $\tilde \alpha$}}
It is apparent in Fig.\ref{phase_space} that the location of the jamming transition depends on the scale of initialization $\tilde \alpha$. From this figure, we observe the following general trend: the jamming transition occurs with less parameters when the test error of the ensemble-averaged predictor is small. It follows the intuition that an easier rule to learn should correspond to less parameters to fit the data.

\paragraph{Effect of the Jamming transition on performance}
As the jamming transition is approached, the norm $\|f_N\|$ of the predictor diverges \cite{Geiger18}, which is responsible in the cusp in the double descent displayed by the test error in Fig.\ref{overfitting}. This divergence was first pointed out for regression \cite{advani2017high}. Yet for classification, the divergence defers quantitatively and is compatible with an inverse power-law, as illustrated in Fig.\ref{doubledescent}.D. It can be understood using an argument \`a la Landau, inspired by results on the perceptron \cite{Franz16, franz2018jamming}. The intuitive idea is that, by increasing the norm of the predictor, one reduces effectively the unit margin required by the hinge loss to fit the data. For $N<N^*$, data cannot be fitted even with a vanishing margin.  

Specifically, consider that the norm of the predictor $\|f_N\|$ is fixed during training and denote by $N^*(\epsilon)$ the jamming transition as a function of the margin $\epsilon$. Assume (as occurs for the perceptron) that $N^*(\epsilon)$ is a smooth function of $\epsilon$, such that $N^*(\epsilon)\approx N^*(0)+ N'^*(0) \epsilon$. For $N$ just above  $N^*(0)$,  margins of magnitude unity cannot be fitted for a fixed norm of the predictor, but they can if that norm is allowed to increase by a factor $ N'^*(0)/(N-N^*(0))$. Indeed it effectively reduces the margin by that amount to some effective value $\tilde \epsilon\equiv \epsilon (N-N^*(0))/N'^*(0)$.  By construction, $\tilde \epsilon$ satisfies $N^*(\tilde\epsilon)=N$. This argument  justifies the observed inverse power-law.

Note that any perturbation to gradient flow (such as early stopping, using stochastic gradient descent or weight decay) will destroy this divergence \footnote{They are relevant perturbations in the sense of critical phenomena.}. Such regularizations are thus most efficient near jamming \cite{Spigler18}.

\section{Double descent and the benefits of overparametrization} 
\label{S3}
To avoid bad minima in the loss landscape, it is thus sufficient to crank up $N$ passed $N^*$. But then, one would naively expect to lower performance and overfit, as illustrated in the right panel of Fig.\ref{curse}.
It is not the case for classification and deep nets, as illustrated in Fig.\ref{overfitting}: performance is very poor and displays a cusp right at the jamming transition point $N^*$, and then continuously improves as $N\rightarrow\infty$! 

A scaling theory can be built to explain quantitatively why performance keeps improving with $N$ passed that point, and how fast the test error reaches its asymptotic behavior \cite{geiger2019scaling}. Essentially, at infinite width learning corresponds to  well-defined algorithms as discussed in introduction, but these algorithms become noisier when the  width is finite. Indeed as $N\sim h^2$ increases, the fluctuations on the learnt output function induced by the random initialization of the weights decrease \cite{neal2018modern}. It follows $\|f_N-\langle f_N\rangle \|\sim N^{-1/4}$ as shown in Fig.\ref{doubledescent}.D, where $\langle f_N\rangle$ is obtained by ensemble-averaging outputs trained with different random initializations. This result is true both for the lazy training \cite{geiger2019scaling} and the feature learning \cite{geiger2019disentangling} regime, as justified below.
The associated increase in test error is quadratic in the fluctuations (it is obvious for a mean square error loss, but is also true for classification under reasonable assumptions \cite{geiger2019scaling}), leading to $\langle \epsilon(f_N)\rangle-\epsilon(\langle f_N\rangle)\sim N^{-1/2}$ as observed \cite{geiger2019scaling}.  This effect is responsible for the double descent, as can be directly checked by considering the training curve $\epsilon(\langle f_N\rangle)$ of the ensemble average function $\langle f_N\rangle$ in Fig.\ref{overfitting} which does not display a second descent. More generally as apparent in Fig.\ref{phase_space}, at fixed $\tilde{\alpha}=\sqrt{h}\alpha$, the ensemble average
test error weakly varies with $h$ once in the over-parametrized regime.

In the lazy training regime, the fluctuations  $\langle \|f_N-\langle f_N\rangle\|\rangle\sim N^{-1/4}$ simply mirror the fluctuations of the NTK induced by initialization, argued to follow $\| \Theta_N-\langle\Theta_N\rangle \|\sim N^{-1/4}\sim h^{-1/2}$ in \cite{geiger2019scaling}, as analytically confirmed since then \cite{hanin2019finite,dyer2019asymptotics}. For simple losses and gradient descent,  the predictor $f_N$ must acquire the same fluctuations \cite{geiger2019scaling}, as observed.

Similar fluctuations induced by initial conditions $\langle \|f_N-\langle f_N\rangle\|\rangle\sim N^{-1/4}$ with identical consequences occur in the feature learning regimes in neural nets \cite{geiger2019disentangling}.
Such fluctuations are certainly expected at initialization for a one-hidden layer, since the density of neurons parameters $\rho$ introduced in Eq.\ref{112} must have finite sampling fluctuations of order $\delta\rho\sim 1/\sqrt{h}\sim N^{-1/4}$ as expected from the central limit theorem. Because in the asymptotic regime $N\rightarrow\infty$ the value of $\rho$ at initialization affects the learnt function, the magnitude of fluctuations $\delta\rho$ (and thus of the learnt function $f_N$) induced by the random initialization will still scale as $N^{-1/4}$ after learning, as rigorously proven for a one-hidden layer in \cite{chen2020dynamical}.  

Overall, this scaling theory supports that for generic data sets and deep architectures, the second descent results from the  noise induced by finite $N$ effects and initialization on the limiting algorithms reached as $N\rightarrow\infty$. One interesting practical implication, apparent in Fig.\ref{phase_space} and Fig.\ref{overfitting}, is that optimal performance is found by ensemble averaging nets of limited width passed their jamming transition.

Since these arguments were proposed, the double descent curve has been analytically computed in the simple case of linear data in the limit of infinite dimension and random features machines \cite{mei2019mean,ghorbani2019linearized}, where it was also shown \cite{d2020double,jacot2020implicit} to  result from the fluctuations of the kernel that vanish with increasing number of neurons, confirming the present explanation.


\begin{figure}[htbp]
   \begin{center}
    \scalebox{0.9}{\import{figures/}{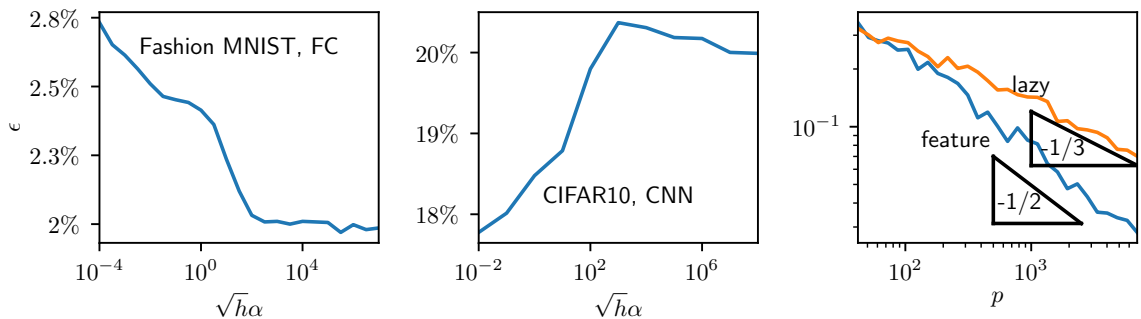}}
     \caption{\underline{Left}, \underline{Center}: Ensemble-average   test error {\it vs.} the scaling parameter  $\tilde{\alpha}=\sqrt{h}\alpha$.  It illustrates that the  NTK regime (large $\sqrt{h}\alpha$) tends to outperform the feature learning regime (small $\sqrt{h}\alpha$) for FC architecture (\underline{Left}) but not in CNNs (\underline{Center}). In the latter case, the training curve exponent $\beta$ is larger in the feature learning (blue curve, $\beta\approx 1/2$) regime than in the NTK one (orange  curve, $\beta\approx 1/3$).
    From \cite{geiger2019disentangling}.}
     \label{comparison}
   \end{center}
\end{figure}

\section{Disentangling the NTK and feature learning regimes}
\label{S4}
As discussed in the previous section, wider nets are more predictive.
A comparable gain of performance can however be obtained already at intermediate width, by ensemble averaging outputs
over multiple initialization of the weights. As apparent in Fig.\ref{phase_space}, the rescaled magnitude of
initialization $\tilde\alpha=\alpha \sqrt{h}$ which allows to cross-over between the lazy training and feature learning regimes does affect performance. 
Which regime best characterises deep nets used in practice? Which one  performs better?
Some predictions of the NTK regime appear to hold in realistic architectures ~\cite{lee2019wide},
and training nets in NTK limit can achieve good performance on real datasets~\cite{arora2019exact,arora2019harnessing,shankar2020neural}.
Yet, in several cases the feature learning regime beats the NTK \cite{woodworth2020kernel,chizat2019lazy}, in line with the common idea that
building an abstract representation of the data such as sketched in Fig.\ref{fig:architecture} is useful. Several theoretical works show that NTK under-performs for specific, simple models of data \cite{chizat2020implicit,yehudai2019power,bach2017breaking,ghorbani2019limitations,ghorbani2020neural}.

In \cite{geiger2019disentangling}, this question was investigated systematically for deep nets in the $(\alpha,h)$ plane, where $\alpha$ is the scale of initialization introduced in \cite{chizat2019lazy}
as defined in Eq.\ref{eq:bachmodel}. This study developed  numerical methods of adaptive learning rates to follow  gradient flow while changing $\alpha$ by ten decades. The main results are as follows:

(i) The cross-over between the two regimes occur when $\alpha\sqrt{h}=O(1)$ as apparent in Fig.\ref{phase_space}, extending the result of \cite{chizat2019lazy} limited to one hidden layer nets. 
For $\alpha\sqrt{h}\ll1$, $\|\Delta \Theta\|/\|\Theta\|\gg1$ while the opposite holds true when $\alpha\sqrt{h}\gg1$. Here $\Delta \Theta=\Theta(t)-\Theta(t=0)$ characterises the evolution of the tangent kernel and $t$ is the learning time. It is convenient to divide the loss by $\alpha^2$ and consider $\tilde {\cal L}\equiv {\cal L}/\alpha^2$, which can be shown to ensure that the dynamics occurs on a time scale independent of $\alpha$ in the large $\alpha$ limit \cite{chizat2019lazy,geiger2019disentangling}. This result holds true for the usual choices of loss (cross-entropy, hinge, etc...) at any finite time. It also holds true for $t\rightarrow\infty$ if the hinge loss is used, since in that case convergence to a zero loss occurs in finite time independently of $h$ and $\alpha$.

Here we provide a schematic argument justifying this result, see \cite{geiger2019disentangling} for a more detailed analysis. The variation of the output $f$ with respect to the pre-activation $a$ (which are  of order one at initialization) of a given neuron is of order $\partial f/\partial a\sim \alpha/\sqrt{h}$. This result is obvious for the last hidden layer (using that the last weights are of order $1/\sqrt{h}$), but can be justified recursively at all layers, as discussed in \cite{geiger2019disentangling} and derived implicitly in the NTK study \cite{jacot2018neural}.
For gradient flow, the variation $\Delta a$ of pre-activation due to the evolution of the bias (considering the previous weights leads to a similar scaling) must be of order $\Delta a\sim t (\partial \tilde{\cal L}/\partial f)( \partial f/\partial a)\sim t/\alpha \sqrt{h}$ using that $\partial \tilde{\cal L}/\partial f\sim 1/\alpha^2$. Thus $\Delta a$ is of order $1/\alpha\sqrt{h}$ at the end of training (since a zero hinge loss is reached   on a time scale $t=O(1)$). The NTK regime must thus break down when $\Delta a\sim 1$ when the relation between weights and output must become non-linear, corresponding to a cross-over for $\alpha^*\sim 1/\sqrt h$ for large $h$.

A similar line of thought can be used at intermediate width. When reducing $h$ so as to approach the jamming transition from the over-parametrized phase, the norm of the output -- and therefore weights and  pre-activations -- explode as reviewed in Section \ref{S2}. One is never then in the lazy training regime, and the relationship between the variation of weights and the output must  become non-linear. Thus, in the vicinity of the jamming  transition, the networks always lie in the feature learning regime.  Consequently, the cross-over lines separating lazy and feature learning must bend up and never cross the jamming line in Fig.\ref{phase_space}. We do observe curves qualitatively bending up as expected (yet it is hard to make precise measurements very close to jamming).

(ii) {\it For fully-connected nets and gradient descent, the NTK  tends to outperform the feature learning regime},  as exemplified in the Left panel of Fig.\ref{comparison}.
This result was found for a variety of data sets (MNIST, Fashion-MNIST, CIFAR10, etc...), except for MNIST 10 PCA. It is apparent in Fig.\ref{phase_space}, where the best performance for MNIST 10 PCA appears to occur the cross-over region $\tilde \alpha\sim 1$. 

(iii) For CNN architectures, feature learning outperforms the NTK regime as shown in the central panel of Fig.\ref{comparison}. It corresponds to a larger training curve exponent $\beta$,
as appears in the right panel of Fig.\ref{comparison}. (ii,iii) were recently confirmed by the Google team \cite{lee2020finite}.

These observations raise various questions. What are the advantages and drawbacks of feature learning, and why are the latter more apparent in FC rather than CNN architectures? For modern CNN architectures, is the improvement of the learning curve exponent $\beta$ in the feature learning regime key to understand how the curse of dimensionality is beaten? Is it associated with learning invariants in the data? In our opinion, the answers to these questions are yet unknown. To start tackling these questions, in the next section we study a simple model of invariant data for which the improvement of the learning curve exponent $\beta$ in the feature learning regime can be computed.


\section{Learning simple invariants by compressing irrelevant input dimensions}
\label{S5}
\paragraph{How can the curse of dimensionality be beaten?} A favorable aspect of various data sets such as images is that the data distribution $P(\bf x)$ is very anisotropic,
consistent with the notion that the data lie in a manifold of lower dimension.
In that case, the distance between neighboring data reduces faster with growing $P$, improving  kernel methods \cite{spigler2019asymptotic}. The positive effect of a moderate anisotropy can  also be shown for kernel classification \cite{paccolat2020isotropic} and regression  \cite{ghorbani2020neural,goldt2019modelling}. 
Yet, even in simple data sets like MNIST of CIFAR10, the intrinsic dimensions remain significant ($d_{int}\approx 14$ and $d_{int}\approx 30$ respectively \cite{spigler2019asymptotic}). This effect helps but cannot resolve the curse of dimensionality for complex data set: indeed using Laplace or Gaussian kernels (which are very similar to the NTK of fully connected nets) lead to $\beta\approx 0.4$ for MNIST (which is decent) but $\beta\approx 0.1$ for CIFAR10, which implies very slow learning as the number of data increases.  

As discussed in the introduction, another popular idea is that data sets as images are learnable because they display many invariant transformations that leave the labels unchanged. By becoming insensitive to those, the network essentially reduces the dimension of the data. 
This view is consistent with  the notion that  deep learning leads to an abstract neural representation of the data sketched in Fig.\ref{fig:architecture}.
 This effect may be responsible for our observation in  the right panel of Fig.\ref{comparison} that the training curve exponent $\beta$ is more favorable in
 the feature learning regime. However, understanding how invariants are built dynamically and how it affects performance and  $\beta$ remains a challenge. 
\begin{figure}[htbp]
   \begin{center}
     \rotatebox{+0}{\resizebox{\textwidth}{!}{\includegraphics{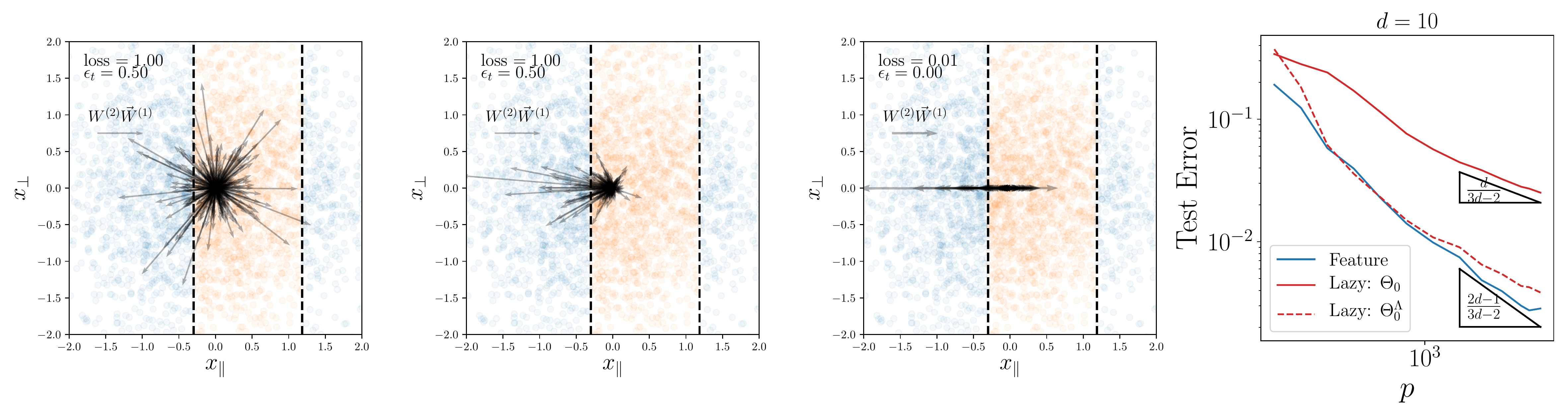}}}
     \caption{\underline{Left}: Distribution of the weights vectors ${\vec W}^{(1)}$ pondered by the neuron second weight $W^{(2)}$ shown at different time points during the learning process (as characterised by the value of the loss in legend) for the stripe model.
     It reveals an alignment of the first layer of weight vectors  in the informative direction.  Training data points are also shown, and coloured depending on their labels. \underline{Right}: Training curve $\epsilon(P)$ of a one hidden layer in the feature learning regime (blue), the NTK regime (red) and the NTK regime after compressing the data in the perpendicular direction by the factor $\Lambda$ (dashed red). Note the similarity between this curve and the blue one, supporting that the main effect of learning feature for these data is the geometric compression of uninformative directions in input space. From \cite{paccolat2020compressing}.}
     \label{compression}
   \end{center}
\end{figure}


Specific models of data can be built \cite{rotskoff2018neural,ongie2019function,ghorbani2019linearized,ghorbani2020neural} in which lazy training does not learn at all, whereas neural nets trained in the feature learning regime succeed. These models however do not capture the two finite and distinct learning curve exponents $\beta$ observed in the two regimes.


\paragraph{The stripe model} In  \cite{chizat2020implicit,paccolat2020compressing,paccolat2020isotropic}, a simple model of invariant is considered. The labels do not depend on $d_\perp=d-d_\parallel$  directions of input space, so that $y(x)=y(x_\parallel)$ where $x=(x_\perp,x_\parallel)$. $x_\bot$ could correspond for example to uninformative pixels near the boundaries of pictures.
For gradient descent,  with {\it the logistic loss},  a one-hidden layer can be shown to correspond to a max-margin classifier in a certain non-Hilbertian space of functions \cite{chizat2020implicit}. Dimension-independent guarantees on performance can then be obtained if the data can be separated after projection in a low dimensional space. The analysis is rigorous and general, but requires  to go to extremely long times not used in practice and does not predict values for $\beta$.

In  \cite{paccolat2020compressing,paccolat2020isotropic}, the hinge loss is considered. In that case, the dynamics stops after a reasonable time.
If the density of data points does not vanish at the interface between labels, and if the latter is sufficiently smooth (e.g. planar or cylindrical), it is found that the test error decays as a power law in both regimes. For the lazy training regime, scaling arguments  inspired by electrostatics lead to $\beta_{Lazy} = d/(3d-2)$. Feature learning performs better, as one find $\beta_{Feature} = (d + \nicefrac{d_\perp}{2}) / (3d - 2)$. The key effect leading to an improvement of the feature learning regime is illustrated  in Fig.\ref{compression} for $d_\parallel=1$, called the stripe model: due to the absence of gradient in the orthogonal direction, the weights only grow along the $d_\parallel$ dimensions. Thus, for a infinitesimal initialization of the weights ($\alpha\rightarrow 0$), the neurons align along the informative coordinate in the data. Accordingly, in relative terms, they  become less sensitive to  $x_\perp$, which merely acts as a source of noise. This denoising is limited by the finite size of the training set. Specifically, it is found that $\Lambda\equiv W_\parallel/W_\perp\sim \sqrt{P}$ where $W_\parallel$ and $W_\perp$ are the characteristic scales of the first layer of weights in the informative and uninformative directions respectively. This effect is equivalent to a geometrical compression of magnitude $\Lambda$ of the input in the uninformative direction. Indeed performing this compression by considering the transformed data $(x_\parallel,x_\perp/\Lambda)$, and learning these in the NTK regime gives very similar performance as learning the original data in the feature regime, see the right panel of Fig.\ref{compression}.

\begin{figure}[htbp]
   \begin{center}
    \includegraphics[width=\textwidth]{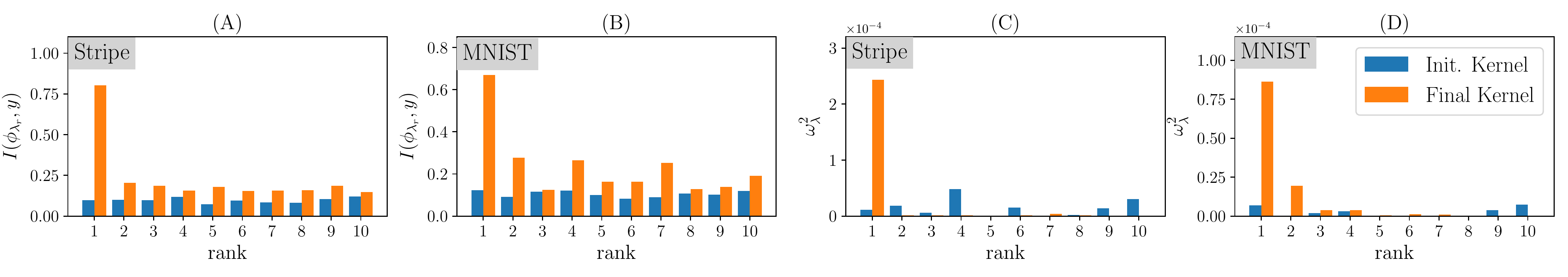}
     \caption{Relationship between the kernel PCA of the NTK (obtained both at initialization and after learning) and the labels. The mutual information (characterizing the amount of correlation) between the top eigenvectors of the Gram matrix and the labels (A, B) and their projection $\omega_\lambda$ (C, D) get large after learning for both the stripe model (learnt with a FC with one hidden-layer) and MNIST (learnt with a CNN). From \cite{paccolat2020compressing}.  
     }
     \label{PCA}
   \end{center}
\end{figure}
\paragraph{Kernel PCA of the NTK reveals a geometric compression of invariants}

The stripe models illustrates that neural nets in the feature learning regime can learn to become insensitive to transformations in the data that do not affect the labels. This insensitivity is limited by the sampling noise associated with a finite data set. Is this phenomenon also occurring for more modern architectures and  more subtle invariants characterizing images?  Measurements of the intrinsic dimension of the neural representation of data \cite{ansuini2019intrinsic,recanatesi2019dimensionality} or mutual information estimates \cite{shwartz2017opening} suggest that it may be so, but as mentioned in introduction these observables have some limitations. 

In the feature learning regime, the NTK evolves in time. This evolution leads to a  better kernel for the task considered. Indeed, using a kernel method based on the NTK obtained at the end of training leads to  essentially identical performance than the neural net itself \cite{paccolat2020compressing}.  This observation is consistent with previous ones showing that the NTK "improves" in time \cite{oymak2019generalization,kopitkov2019neural}, in the following sense. For kernels we can always write $\Theta(x,y)=\psi(x)\cdot\psi(y)$, where $\psi(x)$ is a vector of features. In the case of the NTK, $\psi(x)$ can be chosen as the gradient of the output $f$ with respect to the weights -- see Eq.\ref{eq:ntkformula}.
Kernels tend to perform better \cite{bordelon2020spectrum} if the vector of labels $\{y(x_i)\}_{i=1...P}$ has large coefficients along the first PCAs of the features vectors $\{\psi(x_i)\}_{i=1...P}$ (an operation called kernel PCA \cite{scholkopf_kernel_1999}). Such an alignment between the first PCAs of the features vectors of the NTK and the vector of labels is observed during learning \cite{oymak2019generalization,kopitkov2019neural}.

Although there is no general theory as to why such an  alignment occurs, the stripe model provides a plausible explanation for this effect.  In that case, this improvement must occur because one evolves from an isotropic kernel to an anisotropic one with diminished sensitivity to uninformative directions $x_\perp$. As a result, the top kernel PCA becomes more informative on the label (Fig.\ref{PCA}.A) on which they project more (Fig.\ref{PCA}.C). The same result is observed for a CNN trained on MNIST as shown in Fig.\ref{PCA}.B,D. Overall, this view support that the improvement of the NTK reveals the geometric compression of uninformative directions in the data.

 \section{Conclusion} 
 
 Don't be afraid of bad minima! Just crank-up the number of parameters 
 until you pass a jamming transition. Beyond that point, bad minima are not encountered, and you can bring the loss to zero. Depending on the network width and on how you scale the value of your function at initialization, deep learning can then behave as a kernel method, or can alternatively learn features.  Simple scaling arguments delimit the corresponding phase diagram, and appear to hold for benchmark data sets. Remarkably, these arguments depend very little on the specific data considered. Their practical implication is that ensemble averaging nets with different initialization past the jamming transition can be an effective procedure, as found by different groups \cite{geiger2019scaling,geiger2019disentangling,lee2020finite}.
 
Taking into account where you are in this phase diagram is arguably key to understand outstanding questions on deep learning.
 It is true for example for the role of stochasticity in the dynamics, which appears to improve performance.
It is natural that using stochastic gradient descent instead of gradient flow will help near jamming, because the noise will regularize the divergence of the predictor norm -- an effect that can already be obtained with early stopping \cite{Spigler18}. Likewise, for repulsive particles temperature regularizes singularity near the jamming transition \cite{degiuli2015theory}. Yet, it would be useful to study its effect on performance for the distinct regions of the phase diagram. It would be particularly interesting to know if one of the main effects of stochasticity in the over-parametrized limit is to push upward the cross-over between the lazy and feature learning regime in Fig.\ref{phase_space}, thus improving performance in CNNs where feature learning outperforms lazy training.
 
Ultimately, the central question left to understand is performance, and how the curse of dimensionality is beaten in deep nets. Even at an empirical level, the idea that invariance toward diffeomorphisms is the central phenomenon behind this success is not established. At the theoretical level, this view as not been combined with the recent improvement of our understanding of learning, which as mostly  focused on fully connected nets. In comparison, the effort to model  how CNNs learn features is more modest, despite that only those architectures tend to work well in practice. Such studies would arguably require to design simple canonical models of data presenting  complex invariants, which are currently scarce.

\subsection*{Acknowledgments}
We acknowledge G. Biroli, S. d'Ascoli, F. Cagnetta, A. Favero, F. Gabriel, C. Hongler, A. Jacot, J. Paccolat, L. Pillaud-Vivien, S. Spigler, L. Sagun, U. Tomasini, K. Tyloo and all members of the PCSL group for discussions. This work was partially supported by the grant from the Simons Foundation (\#454953 Matthieu Wyart). M.W. thanks the Swiss National Science Foundation for support under Grant No.~200021-165509.



\pagebreak

\bibliography{main}{}
\bibliographystyle{unsrt}

\end{document}

%% file: figures/fig2A.pdf_tex
\begingroup%
  \makeatletter%
  \providecommand\color[2][]{%
    \errmessage{(Inkscape) Color is used for the text in Inkscape, but the package 'color.sty' is not loaded}%
    \renewcommand\color[2][]{}%
  }%
  \providecommand\transparent[1]{%
    \errmessage{(Inkscape) Transparency is used (non-zero) for the text in Inkscape, but the package 'transparent.sty' is not loaded}%
    \renewcommand\transparent[1]{}%
  }%
  \providecommand\rotatebox[2]{#2}%
  \newcommand*\fsize{\dimexpr\f@size pt\relax}%
  \newcommand*\lineheight[1]{\fontsize{\fsize}{#1\fsize}\selectfont}%
  \ifx\svgwidth\undefined%
    \setlength{\unitlength}{208.44769095bp}%
    \ifx\svgscale\undefined%
      \relax%
    \else%
      \setlength{\unitlength}{\unitlength * \real{\svgscale}}%
    \fi%
  \else%
    \setlength{\unitlength}{\svgwidth}%
  \fi%
  \global\let\svgwidth\undefined%
  \global\let\svgscale\undefined%
  \makeatother%
  \begin{picture}(1,0.84187433)%
    \lineheight{1}%
    \setlength\tabcolsep{0pt}%
    \put(0,0){\includegraphics[width=\unitlength,page=1]{fig2A.pdf}}%
  \end{picture}%
\endgroup%

%% file: figures/fig2B.pdf_tex
\begingroup%
  \makeatletter%
  \providecommand\color[2][]{%
    \errmessage{(Inkscape) Color is used for the text in Inkscape, but the package 'color.sty' is not loaded}%
    \renewcommand\color[2][]{}%
  }%
  \providecommand\transparent[1]{%
    \errmessage{(Inkscape) Transparency is used (non-zero) for the text in Inkscape, but the package 'transparent.sty' is not loaded}%
    \renewcommand\transparent[1]{}%
  }%
  \providecommand\rotatebox[2]{#2}%
  \newcommand*\fsize{\dimexpr\f@size pt\relax}%
  \newcommand*\lineheight[1]{\fontsize{\fsize}{#1\fsize}\selectfont}%
  \ifx\svgwidth\undefined%
    \setlength{\unitlength}{208.43988374bp}%
    \ifx\svgscale\undefined%
      \relax%
    \else%
      \setlength{\unitlength}{\unitlength * \real{\svgscale}}%
    \fi%
  \else%
    \setlength{\unitlength}{\svgwidth}%
  \fi%
  \global\let\svgwidth\undefined%
  \global\let\svgscale\undefined%
  \makeatother%
  \begin{picture}(1,0.84190586)%
    \lineheight{1}%
    \setlength\tabcolsep{0pt}%
    \put(0,0){\includegraphics[width=\unitlength,page=1]{fig2B.pdf}}%
  \end{picture}%
\endgroup%

%% file: figures/fig2C.pdf_tex
\begingroup%
  \makeatletter%
  \providecommand\color[2][]{%
    \errmessage{(Inkscape) Color is used for the text in Inkscape, but the package 'color.sty' is not loaded}%
    \renewcommand\color[2][]{}%
  }%
  \providecommand\transparent[1]{%
    \errmessage{(Inkscape) Transparency is used (non-zero) for the text in Inkscape, but the package 'transparent.sty' is not loaded}%
    \renewcommand\transparent[1]{}%
  }%
  \providecommand\rotatebox[2]{#2}%
  \newcommand*\fsize{\dimexpr\f@size pt\relax}%
  \newcommand*\lineheight[1]{\fontsize{\fsize}{#1\fsize}\selectfont}%
  \ifx\svgwidth\undefined%
    \setlength{\unitlength}{206.90233846bp}%
    \ifx\svgscale\undefined%
      \relax%
    \else%
      \setlength{\unitlength}{\unitlength * \real{\svgscale}}%
    \fi%
  \else%
    \setlength{\unitlength}{\svgwidth}%
  \fi%
  \global\let\svgwidth\undefined%
  \global\let\svgscale\undefined%
  \makeatother%
  \begin{picture}(1,0.83837109)%
    \lineheight{1}%
    \setlength\tabcolsep{0pt}%
    \put(0,0){\includegraphics[width=\unitlength,page=1]{fig2C.pdf}}%
  \end{picture}%
\endgroup%

%% file: figures/fig2D.pdf_tex
\begingroup%
  \makeatletter%
  \providecommand\color[2][]{%
    \errmessage{(Inkscape) Color is used for the text in Inkscape, but the package 'color.sty' is not loaded}%
    \renewcommand\color[2][]{}%
  }%
  \providecommand\transparent[1]{%
    \errmessage{(Inkscape) Transparency is used (non-zero) for the text in Inkscape, but the package 'transparent.sty' is not loaded}%
    \renewcommand\transparent[1]{}%
  }%
  \providecommand\rotatebox[2]{#2}%
  \newcommand*\fsize{\dimexpr\f@size pt\relax}%
  \newcommand*\lineheight[1]{\fontsize{\fsize}{#1\fsize}\selectfont}%
  \ifx\svgwidth\undefined%
    \setlength{\unitlength}{207.92577548bp}%
    \ifx\svgscale\undefined%
      \relax%
    \else%
      \setlength{\unitlength}{\unitlength * \real{\svgscale}}%
    \fi%
  \else%
    \setlength{\unitlength}{\svgwidth}%
  \fi%
  \global\let\svgwidth\undefined%
  \global\let\svgscale\undefined%
  \makeatother%
  \begin{picture}(1,0.83672431)%
    \lineheight{1}%
    \setlength\tabcolsep{0pt}%
    \put(0,0){\includegraphics[width=\unitlength,page=1]{fig2D.pdf}}%
  \end{picture}%
\endgroup%

%% file: figures/fig2E.pdf_tex
\begingroup%
  \makeatletter%
  \providecommand\color[2][]{%
    \errmessage{(Inkscape) Color is used for the text in Inkscape, but the package 'color.sty' is not loaded}%
    \renewcommand\color[2][]{}%
  }%
  \providecommand\transparent[1]{%
    \errmessage{(Inkscape) Transparency is used (non-zero) for the text in Inkscape, but the package 'transparent.sty' is not loaded}%
    \renewcommand\transparent[1]{}%
  }%
  \providecommand\rotatebox[2]{#2}%
  \newcommand*\fsize{\dimexpr\f@size pt\relax}%
  \newcommand*\lineheight[1]{\fontsize{\fsize}{#1\fsize}\selectfont}%
  \ifx\svgwidth\undefined%
    \setlength{\unitlength}{223.19151631bp}%
    \ifx\svgscale\undefined%
      \relax%
    \else%
      \setlength{\unitlength}{\unitlength * \real{\svgscale}}%
    \fi%
  \else%
    \setlength{\unitlength}{\svgwidth}%
  \fi%
  \global\let\svgwidth\undefined%
  \global\let\svgscale\undefined%
  \makeatother%
  \begin{picture}(1,0.73637022)%
    \lineheight{1}%
    \setlength\tabcolsep{0pt}%
    \put(0,0){\includegraphics[width=\unitlength,page=1]{fig2E.pdf}}%
  \end{picture}%
\endgroup%

%% file: figures/fig2F.pdf_tex
\begingroup%
  \makeatletter%
  \providecommand\color[2][]{%
    \errmessage{(Inkscape) Color is used for the text in Inkscape, but the package 'color.sty' is not loaded}%
    \renewcommand\color[2][]{}%
  }%
  \providecommand\transparent[1]{%
    \errmessage{(Inkscape) Transparency is used (non-zero) for the text in Inkscape, but the package 'transparent.sty' is not loaded}%
    \renewcommand\transparent[1]{}%
  }%
  \providecommand\rotatebox[2]{#2}%
  \newcommand*\fsize{\dimexpr\f@size pt\relax}%
  \newcommand*\lineheight[1]{\fontsize{\fsize}{#1\fsize}\selectfont}%
  \ifx\svgwidth\undefined%
    \setlength{\unitlength}{231.62499375bp}%
    \ifx\svgscale\undefined%
      \relax%
    \else%
      \setlength{\unitlength}{\unitlength * \real{\svgscale}}%
    \fi%
  \else%
    \setlength{\unitlength}{\svgwidth}%
  \fi%
  \global\let\svgwidth\undefined%
  \global\let\svgscale\undefined%
  \makeatother%
  \begin{picture}(1,0.6768423)%
    \lineheight{1}%
    \setlength\tabcolsep{0pt}%
    \put(0,0){\includegraphics[width=\unitlength,page=1]{fig2F.pdf}}%
  \end{picture}%
\endgroup%

%% file: figures/fig2G.pdf_tex
\begingroup%
  \makeatletter%
  \providecommand\color[2][]{%
    \errmessage{(Inkscape) Color is used for the text in Inkscape, but the package 'color.sty' is not loaded}%
    \renewcommand\color[2][]{}%
  }%
  \providecommand\transparent[1]{%
    \errmessage{(Inkscape) Transparency is used (non-zero) for the text in Inkscape, but the package 'transparent.sty' is not loaded}%
    \renewcommand\transparent[1]{}%
  }%
  \providecommand\rotatebox[2]{#2}%
  \newcommand*\fsize{\dimexpr\f@size pt\relax}%
  \newcommand*\lineheight[1]{\fontsize{\fsize}{#1\fsize}\selectfont}%
  \ifx\svgwidth\undefined%
    \setlength{\unitlength}{231.01629062bp}%
    \ifx\svgscale\undefined%
      \relax%
    \else%
      \setlength{\unitlength}{\unitlength * \real{\svgscale}}%
    \fi%
  \else%
    \setlength{\unitlength}{\svgwidth}%
  \fi%
  \global\let\svgwidth\undefined%
  \global\let\svgscale\undefined%
  \makeatother%
  \begin{picture}(1,0.73768813)%
    \lineheight{1}%
    \setlength\tabcolsep{0pt}%
    \put(0,0){\includegraphics[width=\unitlength,page=1]{fig2G.pdf}}%
  \end{picture}%
\endgroup%

%% file: figures/fig2H.pdf_tex
\begingroup%
  \makeatletter%
  \providecommand\color[2][]{%
    \errmessage{(Inkscape) Color is used for the text in Inkscape, but the package 'color.sty' is not loaded}%
    \renewcommand\color[2][]{}%
  }%
  \providecommand\transparent[1]{%
    \errmessage{(Inkscape) Transparency is used (non-zero) for the text in Inkscape, but the package 'transparent.sty' is not loaded}%
    \renewcommand\transparent[1]{}%
  }%
  \providecommand\rotatebox[2]{#2}%
  \newcommand*\fsize{\dimexpr\f@size pt\relax}%
  \newcommand*\lineheight[1]{\fontsize{\fsize}{#1\fsize}\selectfont}%
  \ifx\svgwidth\undefined%
    \setlength{\unitlength}{232.61575293bp}%
    \ifx\svgscale\undefined%
      \relax%
    \else%
      \setlength{\unitlength}{\unitlength * \real{\svgscale}}%
    \fi%
  \else%
    \setlength{\unitlength}{\svgwidth}%
  \fi%
  \global\let\svgwidth\undefined%
  \global\let\svgscale\undefined%
  \makeatother%
  \begin{picture}(1,0.7326158)%
    \lineheight{1}%
    \setlength\tabcolsep{0pt}%
    \put(0,0){\includegraphics[width=\unitlength,page=1]{fig2H.pdf}}%
  \end{picture}%
\endgroup%